\documentclass{amsart}

\usepackage{amsmath}
\usepackage{amssymb} 
\usepackage{amsthm,amsbsy}
\usepackage{url}            
\usepackage{amsfonts}       
\usepackage{microtype}      
\usepackage{lipsum}
\usepackage{color}
\usepackage[linesnumbered,ruled,vlined]{algorithm2e}
\SetKwInput{KwInput}{Input}                
\SetKwInput{KwOutput}{Output}              
\usepackage{bbm}

\newtheorem{lemma}{Lemma}
\newtheorem{prop}{Proposition}
\newtheorem{remark}{Remark}
\usepackage{graphicx} 
\usepackage{multirow}
\usepackage{float,lscape}

\title[Large-scale SSL via Graph Structure Learning over High-dense Points]{Large-Scale Semi-supervised Learning via Graph Structure Learning over High-dense Points}
\author[Z. Wang]{Zitong Wang}
\author[L. Wang] {Li Wang}
\author[R. Chan]{Raymond Chan}
\author[T. Zeng]{Tieyong Zeng}
\thanks{
		Zitong Wang is with the Department
		of Mathematics, The Chinese University of Hong Kong. 
		E-mail:  ztwang@math.cuhk.edu.hk. \\
		\indent Li Wang is with the Department
		of Mathematics and Department of Computer Science and Engineering, University of Texas at Arlington, Texas, 76019 USA. E-mail: li.wang@uta.edu. Corresponding Author. \\
		\indent Raymond Chan is with the Department
		of Mathematics, City University of Hong Kong, Hong Kong. E-mail: rchan.sci@cityu.edu.hk.\\
		\indent Tieyong Zeng is with the Department
		of Mathematics, The Chinese University of Hong Kong, Hong Kong. E-mail: zeng@math.cuhk.edu.hk. 
	}

\begin{document}
\maketitle

\begin{abstract}
We focus on developing a novel scalable graph-based semi-supervised learning (SSL) method for a small number of labeled data and a large amount of unlabeled data. 
Due to the lack of labeled data and the availability of large-scale unlabeled data, existing SSL methods usually encounter either suboptimal performance because of an improper graph or the high computational complexity of the large-scale optimization problem. In this paper, we propose to address both challenging problems by constructing a proper graph for graph-based SSL methods. Different from existing approaches, we simultaneously learn a small set of vertexes to characterize the high-dense regions of the input data and a graph to depict the relationships among these vertexes. A novel approach is then proposed to construct the graph of the input data from the learned graph of a small number of vertexes with some preferred properties. Without explicitly calculating the constructed graph of inputs, two transductive graph-based SSL approaches are presented with the computational complexity in linear with the number of  input data. Extensive experiments on synthetic data and real datasets of varied sizes demonstrate that the proposed method is not only scalable for large-scale data, but also achieve good classification performance, especially for extremely small number of labels.
\end{abstract}

\section{Introduction}
Semi-supervised learning is an important learning paradigm for the situations that a large amount of data are easily obtained, but only a few labeled data are available due to the laborious or expensive annotation process \cite{semilearningbook}. A variety of SSL methods have been proposed over the past decades, such as transductive support vector machines (SVM) \cite{joachims1999transductive,collobert2006large}, co-training \cite{blum1998combining}, generative model \cite{nigam2000text}, and graph-based SSL \cite{liu2010large,zhu2003semi,belkin2006manifold,Zhou:2003}. Among them, graph-based SSL methods attract wide attention due to their superior performance including manifold regularization \cite{belkin2006manifold} and label propagation \cite{Zhou:2003}. However, these methods usually suffer from the high computational complexity of computing kernel matrix or graph Laplacian matrix and the optimization problem on a large amount of optimized variables. Moreover, the quality of the input graph becomes critically important for graph-based SSL methods.

Most existing SSL methods construct either a dense matrix from a prefixed kernel function \cite{Zhou:2003} or a sparse matrix from a neighborhood graph \cite{belkin2006manifold}. These graph construction approaches have drawbacks. The dense matrix is computationally impractical for large-scale data with high storage requirement, and the neighborhood graph is less robust for data with varied density regions.
Graphs that are commonly used in graph-based methods are the $K$-nearest neighbor ($K$-NN) graph and the $\epsilon$-neighborhood graph \cite{belkin2006manifold}. Dramatic influences of these two graphs on clustering techniques have been studied in \cite{Maier2009}. The $\epsilon$-neighborhood graph could result in disconnected components or subgraphs in the dataset or even isolated singleton vertices. The $b$-matching method is applied to learn a better $b$-nearest neighbor graph via loopy belief propagation \cite{Jebara2009}, but it is hard to scale up to large-scale datasets. As stated in \cite{elhamifar2011sparse}, it is improper to use a fixed neighborhood size since the curvature of manifold and the density of data points may be different in regions of the manifold. As a result, it is less reliable to directly construct the $K$-NN graph in a high-dimensional space as the input to the graph-based SSL methods. 

In addition to the quality of the constructed graph, the scalability is another important issue of graph-based SSL methods. Various methods have been proposed to solve this issue by concentrating on either learning an efficient representation of a graph or developing the scalable optimization methods. A nonparametric inductive function is proposed to do label prediction based on a subset of samples \cite{delalleau2005efficient}. However, this method loses considerable information by ignoring the topology structure within the majority part of the input data. Nystrom-approximation of the graph adjacency matrix is proposed in  \cite{zhang2009prototype}, while the approximated graph Laplacian matrix is not guaranteed to be positive semi-definite. Moreover, random projections including Nystrom method and random features are incorporated into the manifold regularization \cite{mcwilliams2013correlated,sivananthan2018manifold}. The smooth eigen-vector of the graph Laplacian matrix calculated by a numerical method is used to specify the label prediction function \cite{fergus2009semi}, but the assumption based on the dimensional-separable data is restrictive in general.
Anchor graph regularization (AGR) \cite{liu2010large} constructs the graph of the input data based on a small set of anchor points that are the centroids of the $k$-means method, and the constructed graph can be explained by the stationary Markov random walks on a bipartite graph. Graph sparsification methods \cite{cesa2013random} are proposed to approximate the graph Laplacian of input data using a spanning tree and the fast labeling methods are then proposed under the online learning setting. 

Another line of research on SSL is to design fast optimization algorithm for solving graph-based SSL problems. The dual problem of the Laplacian SVM under the manifold regularization \cite{belkin2006manifold} subjects to a sparsity constraint is formulated as a center-constrained minimum
enclosing ball problem \cite{tsang2007large}, which can be efficiently solved by the core vector machine \cite{tsang2005core}. The primal problem of the Laplacian SVM is also solved by the preconditioned conjugate descend \cite{melacci2011laplacian}, which allows to compute the approximation solutions
with roughly the same classification accuracy as the optimal ones, considerably reducing the training time. Distributed approaches have also been explored by decomposing the large-scale problem into smaller ones \cite{chang2017distributed}.

In this paper, we focus on both the quality of the graph constructed from the input data and the scalability of the graph-based SSL methods. We take a strategy that similar to AGR, where the graph or affinity matrix over the input data is implicitly represented by a subset of points. And then, the efficient graph-based SSL methods are designed to take the advantage of the graph representation for large-scale data. The main contributions of this paper are summarized as follows:
\begin{itemize}
	\item A subset of representative points are learned from the input data, each of which governs a nearby part of high-dense regions. Moreover, the graph structure such as a spanning tree is learned simultaneously to capture the similarity among these high-dense points. Our label predication model for SSL recovers the graph structure learning in the unsupervised setting \cite{mao2016principal} with an alternative explanation from the perspective of density modeling.
	
	\item A novel graph construction approach is proposed to take the advantage of both the assignment of each input data to its high-dense points and the graph over these high-dense points. We prove that given the set of anchor points, the graph construction in AGR is a special case of our proposed approach. Moreover, the constructed graph is symmetric, and the graph Laplacian matrix is guaranteed to be positive semi-definite. We also show the spectrum properties of the constructed graph via the convergence property of matrix series.
	
	\item We demonstrate that our constructed graph can be efficiently incorporated by two variants of graph-based SSL methods including the approach used in local and global consistency (LGC) \cite{Zhou:2003} and its variant of learning a linear prediction function in AGR \cite{liu2010large}. We show that both SSL methods have linear computation complexity with the number of input data.
	
	\item Extensive experiments are conducted. We demonstrate the advantages of our proposed SSL method in details by using a synthetic data. We then evaluate our methods on various input datasets of different sizes for both the classification performance of inferring the unlabeled data and the scalability. Our experimental results show that our methods not only achieve competitive performance to baseline methods but also are more efficient for large-scale data. We show that our methods outperform baselines significantly for datasets with extremely small number of labels.
	
\end{itemize}

The rest of the paper is organized as follows. We review two most related work in Section \ref{sec:related-work}. In Section \ref{sec:proposed-method}, we present the proposed method in details. Extensive experiments are conducted in Section \ref{sec:experiments}. We conclude this work in Section
\ref{sec:conclusions}.

\section{Related Work} \label{sec:related-work}
We discuss two graph-based SSL methods that are considered as the basis of our proposed work. The first work is the regularization framework based on LGC \cite{Zhou:2003}, and the other is AGR based on large-scale anchor graph construction \cite{liu2010large}. 

Given a dataset $X=[\mathbf{x}_1,\ldots,\mathbf{x}_n] \in \mathbb{R}^{d \times n}$ and a label set $\mathcal{Y} = \{ 1, \ldots, c \}$, the first $l$ data have labels denoted by $\{ (\mathbf{x}_i, y_i) \}_{i=1}^l$ where $y_i \in \mathcal{Y}$, and the rest of data are unlabeled. Let $\|\mathbf{x}\|$ be the $2$-norm of vector $\mathbf{x}$, $\|X\|_{\text{fro}}$ be the Frobenius norm of matrix $X$, $\textbf{trace}(X)$ be the trace of matrix $X$, and $\textbf{1}_k$ is the length $k$ vector of all ones.

LGC \cite{Zhou:2003} builds on the following two assumptions:
\begin{itemize}
	\item \textbf{Local assumption:} nearby points are likely to have the same label;
	
	\item \textbf{Global assumption:} points on the same structure (e.g., clusters or manifold) are likely to have the same label.
\end{itemize}
Let $F \in \mathbb{R}^{n \times c}$ be the classification matrix of $X$ where the label of the $i$th data point can be obtained by

\begin{small}\vspace{-0.1in}
	\begin{align}
	y_i = \arg \max_{j \in\{1,\cdots,c \} } F_{i,j}, \forall i=1,\ldots,n. \label{eq:decision}
	\end{align}	
\end{small}\noindent
Accordingly, the class labels of $X$ can be encoded as $Y \in [0,1]^{n \times c}$ where $Y_{i,j}=1$ if $\mathbf{x}_i$ is labeled with $y_i=j$, and $Y_{i,j}=0$ otherwise. $Y_{i,j}=0, \forall j=1,\ldots,c$ for all unlabeled data, there is no bias to any specific label for unlabeled data. 

Let $F_i$ be the $i$th row of matrix $F$, and $I_n$ be an $n$ by $n$ identity matrix. Mathematically, to formulate the local assumption, a non-negative symmetric matrix $W \in \mathbb{R}_+^{n \times n}$ with diagonal elements as zeros is introduced, and the optimal $F$ that satisfies the local assumption is transformed to minimize the following objective function

\begin{small}\vspace{-0.1in}
	\begin{align}
	L_W(F) &= \frac{1}{2} \sum_{i,j=1}^n W_{ij} \left|\left| \frac{1}{\sqrt{D_{ii}}} F_i - \frac{1}{\sqrt{D_{jj}}} F_j \right|\right|^2_{\text{fro}} \label{eq:local}\\
	&=\textbf{trace}(F^T (I_n-S) F) \nonumber
	\end{align}	
\end{small}\noindent
where $D$ is a diagonal matrix with $D_{ii} = \sum_{j=1}^n W_{ij}, \forall i=1,\ldots,n $ and $S=D^{-1/2} W D^{-1/2}$. Note that $I_n-S$ is a normalized graph Laplacian matrix over $W$. By minimizing $L_W(F) $ with respect to $F$, it imposes that $\mathbf{x}_i$ and $\mathbf{x}_j$ should have the same label according to (\ref{eq:decision}) since $\frac{1}{\sqrt{D_{ii}}} F_i \approx \frac{1}{\sqrt{D_{jj}}} F_j $ if $W_{i,j}$ is large. Hence, nearby points $\mathbf{x}_i$ and $\mathbf{x}_j$ measured by a large $W_{ij}$ should have the same label that can be enforced by minimizing (\ref{eq:local}).

On the other hand, the global assumption is formulated by a square loss given by

\begin{small}\vspace{-0.1in}
	\begin{align}
	L_Y(F) = ||F - Y||_{\text{fro}}^2 \label{eq:global}
	\end{align}	
\end{small}\noindent
which means a good classifier should not change too much from the initial label assignment.  

LGC is proposed to solve 

\begin{small}\vspace{-0.1in}
	\begin{align}
	\min_F L_W(F) + \mu L_Y(F) \label{op:lgc}
	\end{align}	
\end{small}\noindent
where $\mu$ is the trade-off between local and global objectives. Problem (\ref{op:lgc}) has the closed form solution

\begin{small}\vspace{-0.1in}
	\begin{align}
	F = \mu ((1+\mu) I_n - S)^{-1} Y. \label{eq:F}
	\end{align}	
\end{small}\noindent
It is worth noting that the analytic solution (\ref{eq:F}) is impractical for data with large $n$ since the affinity matrix $W$ needs $O(n^2)$ storage and the computation complexity to calculate the inverse of an $n$ by $n$ matrix is $O(n^3)$. To solve these issues, the anchor graph method \cite{liu2010large} is proposed by constructing $W$ from a stochastic matrix $\widehat{Z} \in \mathbb{R}^{n \times k}$, where $k$ is the number of anchor points and $k \ll n$. 

Let $\{\mathbf{u}_1,\ldots,\mathbf{u}_k \}\subset \mathbb{R}^d$ be a set of anchor points, which are the centroids of the $k$-means method over $X$ with the number of clusters as $k$. Moreover, define the $\hat{s}$-nearest neighbors of $\mathbf{x}_i$ from the anchor points as $\mathcal{N}_i$. Two approaches are used to obtain $\widehat{Z}$. One is from the Nadaraya-Watson kernel regression \cite{hastie2005elements} with Gaussian kernel function defined as

\begin{small}\vspace{-0.1in}
	\begin{align}
	\widehat{Z}_{i,r} = \frac{K_h(\mathbf{x}_i, \mathbf{u}_r)}{\sum_{r' \in \mathcal{N}_i} K_h(\mathbf{x}_i, \mathbf{u}_{r'}) }, \forall r \in \mathcal{N}_i, \label{eq:Z-NWK}
	\end{align}	
\end{small}\noindent
where the Gaussian kernel $K_h(\mathbf{x}_i, \mathbf{u}_r) = \exp(-||\mathbf{x}_i - \mathbf{u}_r||/2h^2)$ with the bandwidth $h$ is adopted. The other is the local anchor embedding (LAE) by solving the following problem

\begin{small}\vspace{-0.1in}
	\begin{align}
	\min_{\widehat{Z}} & \frac{1}{2} \sum_{i=1}^n \left|\left|\mathbf{x}_i - \sum_{j\in \mathcal{N}_i} \widehat{Z}_{i,j} \mathbf{u}_j\right|\right|^2 \label{agr:Z}\\
	\textrm{s.t.} & \sum_{j \in \mathcal{N}_i} \widehat{Z}_{i,j}=1, \widehat{Z}_{i,j} \geq 0, \forall i, j\in \mathcal{N}_i, \label{con:simplex}\\
	& \widehat{Z}_{i,j}=0,\forall i, j \not\in \mathcal{N}_i. \label{con:simplex:2}
	\end{align}	
\end{small}\noindent
This problem can be decomposed into $n$ optimization subproblems, and the projected gradient descent method can be used to solve each subproblem efficiently. Note that the simplex constraints (\ref{con:simplex}) can promote sparsity of $\widehat{Z}_{i, j}$ with $j \in \mathcal{N}_i$. Hence, the final $\widehat{Z}$ can be as sparse as the $\hat{s}$-nearest neighbor graph. The basic assumption is that any data point $\mathbf{x}_i$ can be represented by a convex combination of its closest $\hat{s}$ anchors, and the coefficients are preserved for the weights in the nonparametric regression.

In \cite{liu2010large}, the affinity matrix $W$ is then constructed from $\widehat{Z}$ by using the following rule:

\begin{small}\vspace{-0.1in}
	\begin{align}
	W = \widehat{Z} \Lambda^{-1} \widehat{Z}^T, \label{eq:anchor_W}
	\end{align}	
\end{small}\noindent
where $\Lambda \in \mathbb{R}^{k\times k}$ is a diagonal matrix with $\Lambda_{jj}=\sum_{i=1}^n \widehat{Z}_{i,j}$. Specifically, we have, 

\begin{small}\vspace{-0.1in}
	\begin{align}
	W_{i,j} = \sum_{r=1}^k \frac{\widehat{Z}_{i,r} \widehat{Z}_{j,r}}{ \sum_{i'=1}^n \widehat{Z}_{i',r} },\ \forall i, j.
	\end{align}
\end{small}\noindent
Note that $\sum_j W_{i,j} = \sum_{r=1}^k \widehat{Z}_{i,r}=1,\ \forall i$, where the second equality holds because of constraints (\ref{con:simplex}) and (\ref{con:simplex:2}). Hence, the graph Laplacian matrix of $W$ is $\textbf{diag}(W\textbf{1}_n)-W = I_n - W$, so it is normalized. Let

\begin{small}\vspace{-0.1in}
	\begin{align}
	p(u_r | \mathbf{x}_i) = \widehat{Z}_{i, r},~~ p(v_j | u_r) =  \frac{\widehat{Z}_{j,r}}{ \sum_{i'=1}^n \widehat{Z}_{i',r} },
	\end{align}	
\end{small}\noindent
then a bipartite graph consists of vertexes $\{ \mathbf{x}_i \}_{i=1}^{n}$ and anchor points $\{ u_r\}_{r=1}^{k}$ is defined as (\ref{eq:anchor_W}).
It is clear that 

\begin{small}\vspace{-0.1in}
	\begin{align}
	p(\mathbf{x}_j | \mathbf{x}_i) = \sum_{r=1}^k p(u_r | \mathbf{x}_i)p(\mathbf{x}_j | u_r) = W_{i,j}.
	\end{align}	
\end{small}\noindent
As a result, $W$ is the transition matrix over the bipartite graph from one vertex to the other. In other words, the anchor graph defines a one-step transition on the bipartite graph via the stationary Markov random walks.

Although the anchor graph has the nice properties such as the normalized graph Laplacian matrix and the random walk probability interpretation, the heuristic construction based on the $k$-means method for anchor points and the strong assumption of linear reconstruction of data points from anchor points can degrade the learning performance from labeled and unlabeled data.

\section{Graph Structure Learning for SSL based on High-dense Points} \label{sec:proposed-method}

To construct a reliable graph structure for large-scale datasets, we first model the manifold structure of data using graph structure where the vertexes are the representatives of the high density regions of the data and the connectivities of the graph are the similarities between two vertexes. 
We then obtain the graph over the input data by simultaneously learn the set of high density points and their connectivities.
Finally, we apply the graph to graph-based SSL methods such as label propagation.

\subsection{High-dense Points Learning} \label{sec:high-dense}
Given data $\{\mathbf{x}_i \}_{i=1}^n$, we seek a set of points that can represent the high density regions of the data. For the ease of reference, we name them as the high-dense points denoted as $\{ \mathbf{c}_s \}_{s=1}^k$. To model the density of data, we employ kernel density estimation (KDE) on $\{ \mathbf{c}_s \}_{s=1}^k$ to approximate the true distribution of data by assuming that the observed data $\{ \mathbf{x}_i \}_{i=1}^n$ is sampled from the true distribution.

The basic idea of KDE involves smoothing each point $\mathbf{c}_s$ by a  kernel function and summing up all these functions together to obtain a final density estimation. A typical choice of the kernel function
is Gaussian, which is defined as $g(\mathbf{v}) = (2 \pi)^{-d/2} \exp(-\frac{1}{2} \mathbf{v}^T \mathbf{v})$, where $d$ is the dimension of the input $\textbf{v}$. Applying KDE to estimate $\mathbf{x}_i$ over high-dense points leads to the following density function,

\begin{small}\vspace{-0.1in}
	\begin{align}
	p(\mathbf{x}_i | \{ \mathbf{c}_s \}_{s=1}^k ) = \frac{(2 \pi)^{-\frac{d} 2}}{k \sigma^{d}} \sum_{s=1}^k  \exp (-\frac{1}{2\sigma^2} ||\mathbf{x}_i-\mathbf c_s||^2 ), \label{eq:kde-prior}
	\end{align} 	
\end{small}\noindent
where $\sigma$ is the bandwidth parameter of the Gaussian kernel function. To obtain the optimal set  $\{ \mathbf{c}_s \}_{s=1}^k$, we can do the maximum likelihood estimation by solving the following maximum optimization:

\begin{small}\vspace{-0.1in}
	\begin{align}
	\{ \mathbf{c}_s^* \}_{s=1}^k := \arg\max\limits_{\{ \mathbf{c}_s \}_{s=1}^k} \log \prod_{i=1}^n p(\mathbf{x}_i | \{ \mathbf{c}_s \}_{s=1}^k ). \label{op:MLE}
	\end{align}	
\end{small}\noindent
We can further simplify its objective function as

\begin{small}\vspace{-0.1in}
	\begin{align}
	f(C) =  \sum_{i=1}^n \log \sum_{s=1}^k \exp (-\frac{1}{2\sigma^2} ||\mathbf{x}_i-\mathbf c_s||^2 ).
	\end{align}
\end{small}\noindent
where the terms independent of $C=[\mathbf{c}_1,\ldots,\mathbf{c}_k] \in \mathbb{R}^{d\times k}$ are dropped because they do not change the optimal solution of problem (\ref{op:MLE}). Let $C^*=[\mathbf{c}^*_1,\ldots,\mathbf{c}^*_k]$. The first order optimality condition of problem (\ref{op:MLE}) is 

\begin{small}\vspace{-0.1in}
	\begin{align}
	\frac{\partial f(C^*)}{\partial \mathbf{c}_s}=0, \forall s=1,\ldots,k.
	\end{align}	
\end{small}\noindent
As a result, we have the following equation, $\forall s$:

\begin{small}\vspace{-0.1in}
	\begin{align}
	\!\!\!\!\sum_{i=1}^n Z_{i,s}(\mathbf{x}_i-\mathbf{c}^*_s) = 0 \Rightarrow \mathbf{c}_s^* = \sum_{i=1}^n \frac{Z_{i,s}}{\sum_{i=1}^n Z_{i,s}} \mathbf{x}_i, \label{eq:fpi}
	\end{align} 	
\end{small}\noindent
where 

\begin{small}\vspace{-0.1in}
	\begin{align}
	Z_{i,s} = \frac{\exp (-\frac{1}{2\sigma^2} ||\mathbf{x}_i-\mathbf c_s^*||^2 )}{\sum_{s=1}^k \exp (-\frac{1}{2\sigma^2} ||\mathbf{x}_i-\mathbf c_s^*||^2 )}, \forall i, s.\label{eq:kde-Z}
	\end{align} 	
\end{small}\noindent
Hence, problem (\ref{op:MLE}) can be solved by fixed point iteration in terms of equation (\ref{eq:fpi}).

It is worth noting that the high-dense points are different from centroids obtained by the $k$-means method. Maximizing (\ref{op:MLE}) with respect to $\mathbf{c}_s$ is an iteration step towards the high density $p(X|\mathbf{c}_s)$ over the input data. Moreover, we find that $Z_{i,s}$ in (\ref{eq:kde-Z}) is analogous to (\ref{eq:Z-NWK}). The key difference is that the high-dense points $\{ \mathbf{c}_s \}_{s=1}^k$ and $Z$ are jointly optimized, while in (\ref{eq:Z-NWK}) anchors are assumed to be given. Also, our $Z$ is different from the one obtained by the anchor graph method in (\ref{agr:Z}), since our method does not depend on the stringent assumption that any input data point can be recovered by a convex combination of $\hat{s}$-nearest neighbors points.

\subsection{Learning Connectivity over High-dense Points}
In section \ref{sec:high-dense}, we are able to obtain a set of high-dense points to represent the high density regions of the input data. In this section, we further propose to model the connectivity of high-dense points via some graph structure. As required by most graph-based method, we enforce the graph structure as a connected graph. To make the structure more general, we explore the spanning tree structure to characterize the connectivity among high-dense points. 

Let $\mathcal{T}=(\mathcal{V}, \mathcal{E})$ be a spanning tree with $\mathcal{V}$ as the vertexes and $\mathcal{E}$ as the set of edges. Given the set of high-dense points  $\{ \mathbf{c}_s \}_{s=1}^k$, we assign each point $\mathbf{c}_s$ to a vertex of the tree $\mathcal{T}$, i.e., $\mathcal{V} = \{ \mathbf{c}_s \}_{s=1}^k$, the number of vertexes in the tree $\mathcal{T}$ is $k$. Let $G \in \{0,1\}^{k\times k}$ be the connectivity matrix where $G_{i,j}=1$ means $\mathbf{c}_i$ and $\mathbf{c}_j$ are connected, and $G_{i,j}=0$ otherwise. $\mathcal{T}$ is an undirected graph, so $G$ is symmetric. The Euclidean distance between two corresponding high-dense points can be used to measure the dissimilarity between two vertexes. The minimum-cost spanning tree is naturally used to form a tree by setting the cost as the dissimilarity of two vertexes. In this case, we always pick the least dissimilar edge and simultaneously impose the connectivity assumption over all high-dense points. By combining the connectivity learning with the high-dense points learning, we propose a joint optimization problem as:

\begin{small}\vspace{-0.1in}
	\begin{align}
	\max_{C, G \in \mathcal{T}} f(C) - \frac{\lambda_1}{4} \sum_{r=1}^k \sum_{s=1}^k G_{r,s} || \mathbf{c}_r - \mathbf{c}_s ||^2 \label{op:MLE+tree}
	\end{align} 	
\end{small}\noindent
where $\lambda_1$ is a parameter to balance the two objectives.


Suppose $G$ is given. To solve problem (\ref{op:MLE+tree}) with variable $C$, similar to problem (\ref{op:MLE}), we can obtain the first order optimality condition over $C$ as 

\begin{small}\vspace{-0.1in}
	\begin{align}\label{sub:C:1}
	\left[ \sum_{i=1}^n Z_{i,1} (\mathbf{x}_i-\mathbf{c}_1), \ldots,\sum_{i=1}^n Z_{i,k} (\mathbf{x}_i-\mathbf{c}_k) \right] - \lambda_1 C L = 0 
	\end{align} 
\end{small}\noindent
where $L = \textbf{diag}(G\textbf{1}_k) - G$ is the graph Laplacian matrix over $G$. We can rewrite equation (\ref{sub:C:1}) in matrix form as:  

\begin{small}\vspace{-0.1in}
	\begin{align}
	C \Xi - X Z + \lambda_1 C L = 0
	\end{align} 	
\end{small}\noindent
where $\Xi \in \mathbb{R}^{k \times k}$ is a diagonal matrix with the ($s,s$) entry $\sum_{i=1}^n Z_{i,s}$, i.e., $\Xi = \textbf{diag}(Z^T \textbf{1}_n)$. Hence, we have the closed-form solution for optimization problem (\ref{op:MLE+tree}) with variable $C$: 

\begin{small}\vspace{-0.1in}
	\begin{align}
	C = X Z (\Xi + \lambda_1 L)^{-1}. \label{eq:C}
	\end{align} 	
\end{small}\noindent
Given $C$, problem (\ref{op:MLE+tree}) with respect to variable $G$ can be efficiently solved by the Kruskal's algorithm \cite{kruskal1956shortest} for finding a minimum-cost spanning tree. Hence, problem (\ref{op:MLE+tree}) can be solved by alternating the Kruskal's algorithm for $G$ given $C$ and the fixed point iteration method for updating $C$ given $G$ until convergence.

It is worth noting that the fixed point iteration for $C$ involving equation (\ref{eq:kde-Z}) and (\ref{eq:C}) can be alternated with the minimal spanning tree problem, that is, problem (\ref{op:MLE+tree}) can be solved by alternating the minimum spanning tree for $G$, (\ref{eq:kde-Z}) for $Z$ and (\ref{eq:C}) for $C$ until convergence. We notice that our alternating approach is the same as the principal graph learning based on reversed graph embedding \cite{mao2016principal}. However, this work is motivated from finding the set of high-dense points for graph-based SSL methods, which is different from the principal graph learning since the learned graph is the key to model the data in \cite{mao2016principal}, while this work aims to recover the graph over the input data using high-dense points $C$, assignment matrix $Z$, and their connectivity $G$.

\begin{figure*}
	\centering
	\begin{tabular}{@{}c@{}c@{}c@{}c@{}}
		\includegraphics[width=0.25\textwidth]{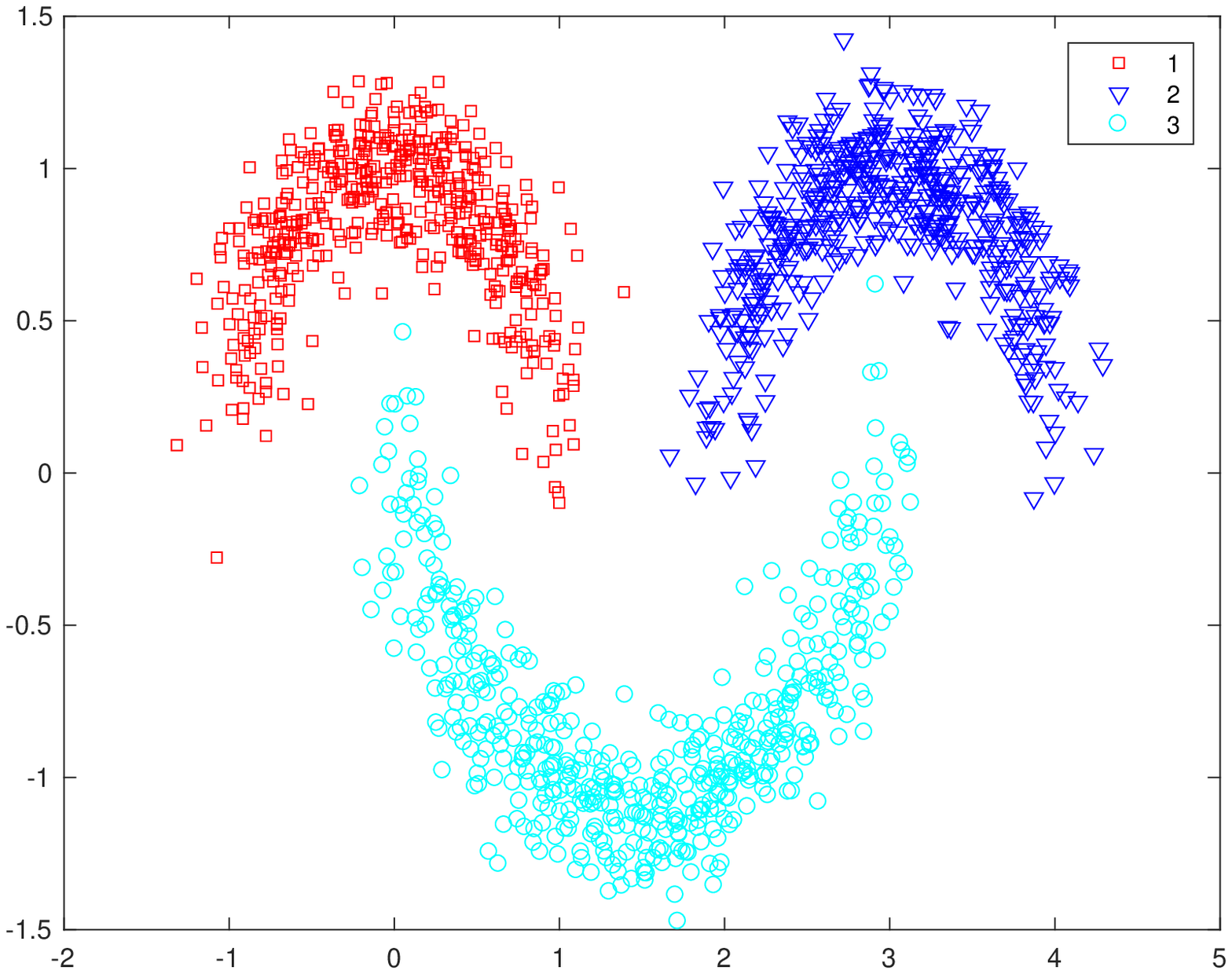}&
		\includegraphics[width=0.25\textwidth]{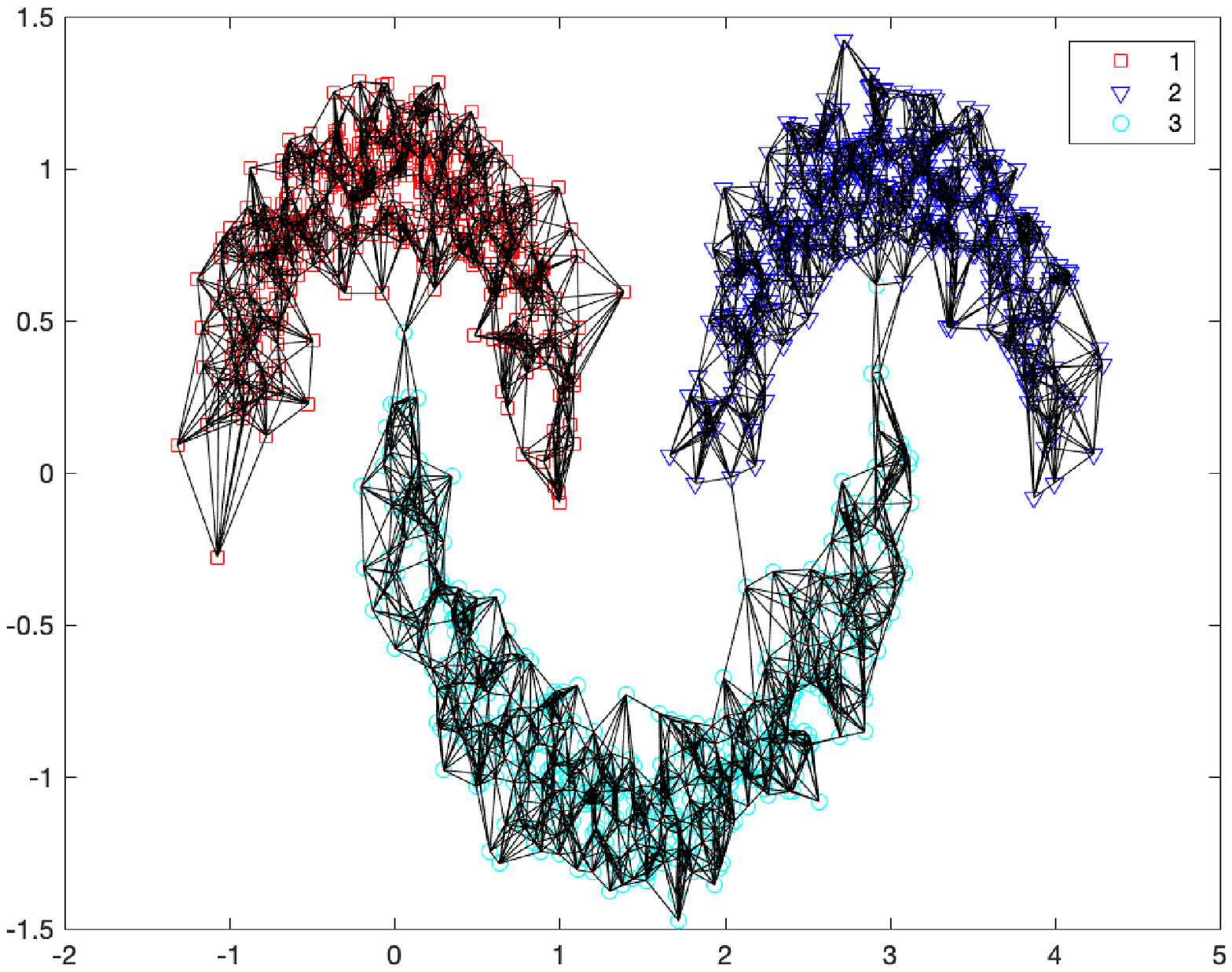}&
		\includegraphics[width=0.25\textwidth]{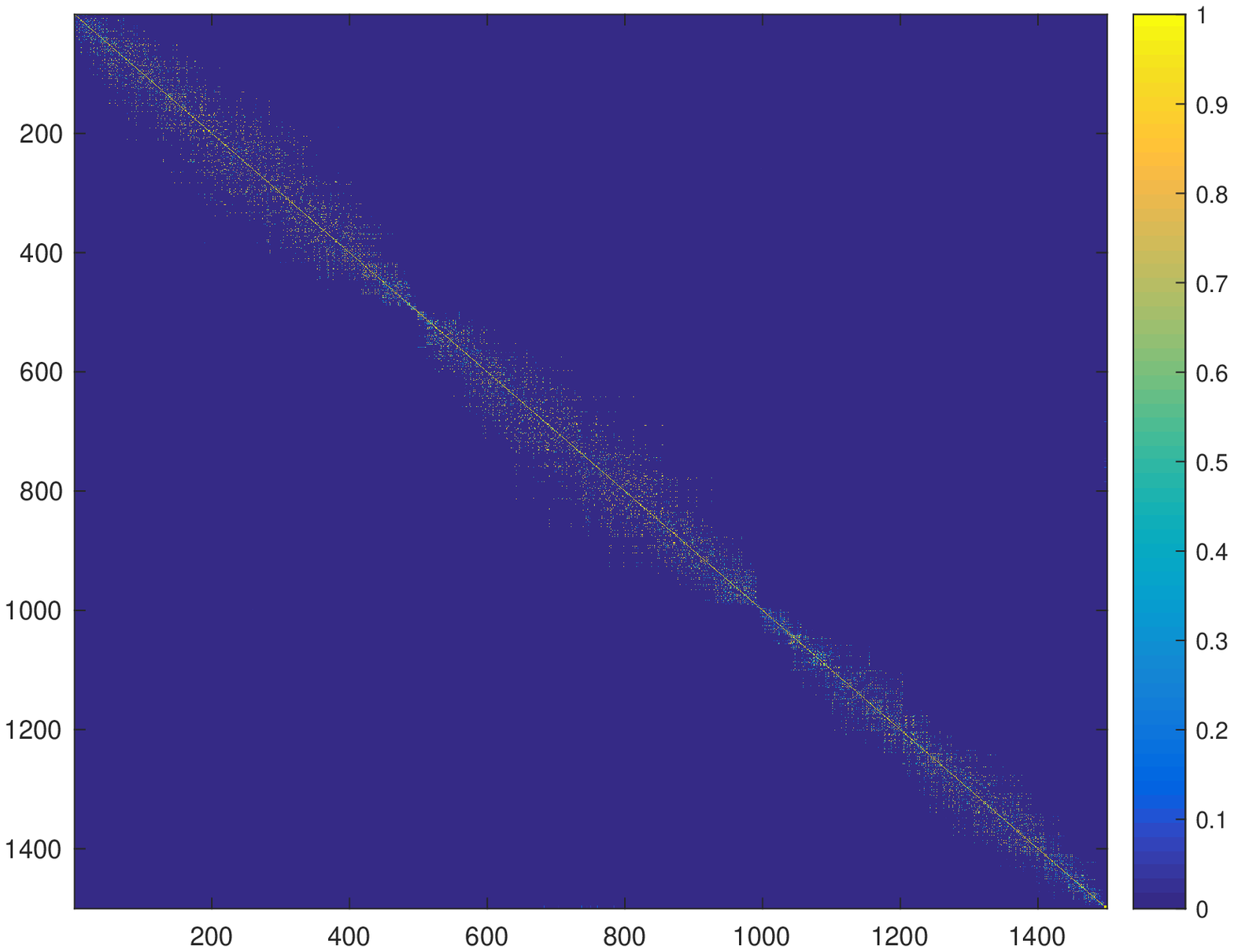}&
		\includegraphics[width=0.25\textwidth]{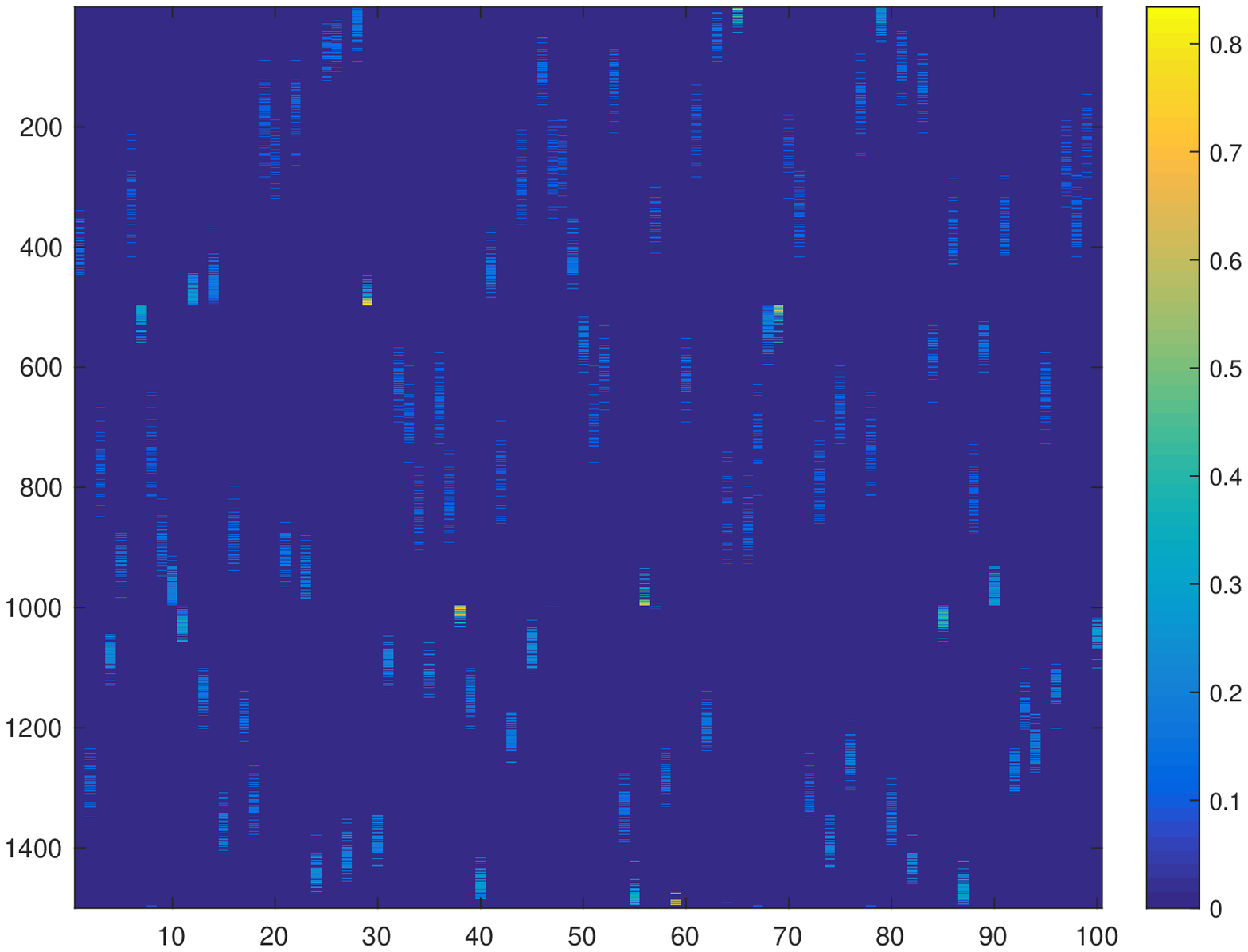}\\
		(a) three-moon data & (b) 10-NN graph & (c) 10-NN affinity matrix & (d) $Z$ of our method \\
		\includegraphics[width=0.25\textwidth]{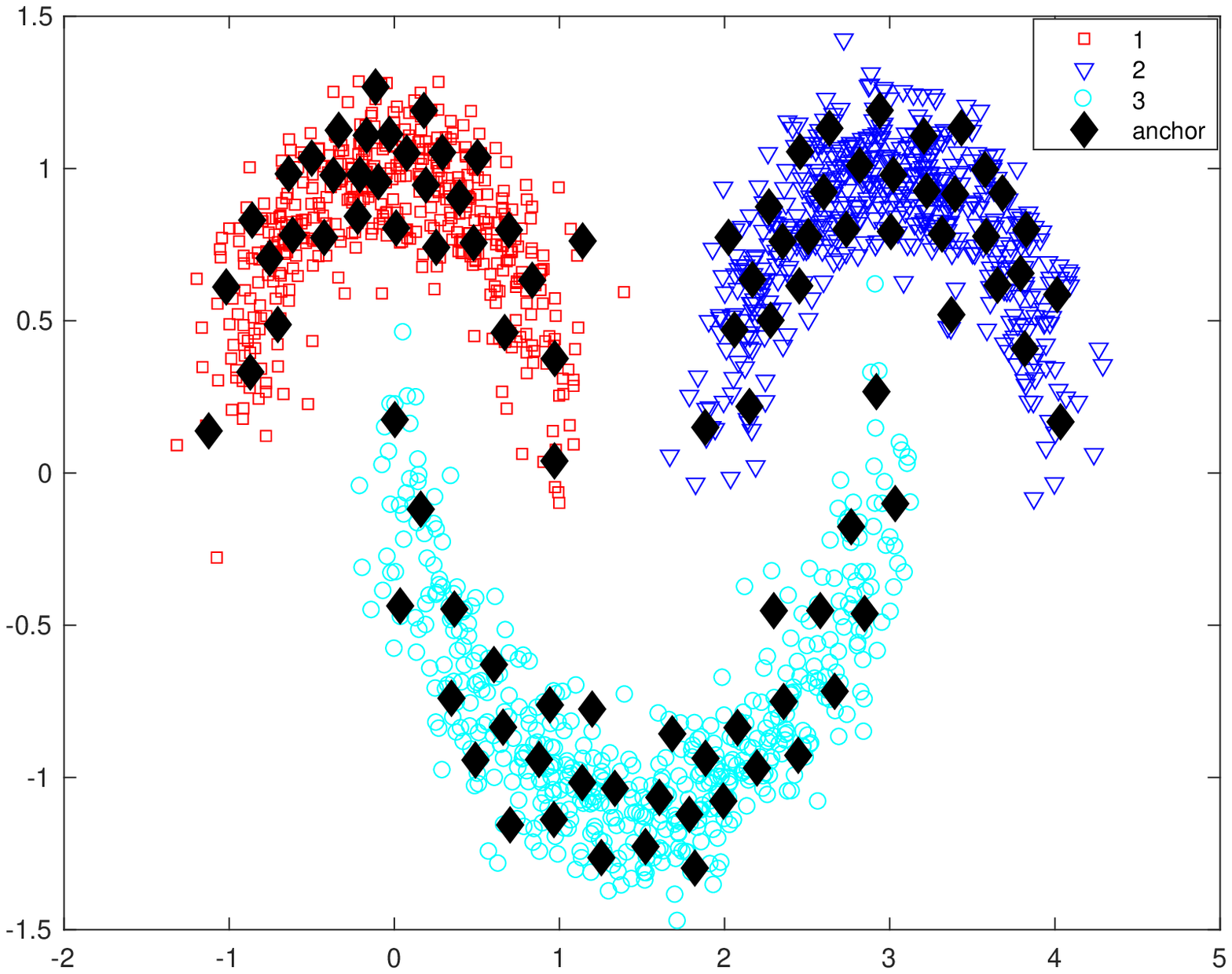}&
		\includegraphics[width=0.25\textwidth]{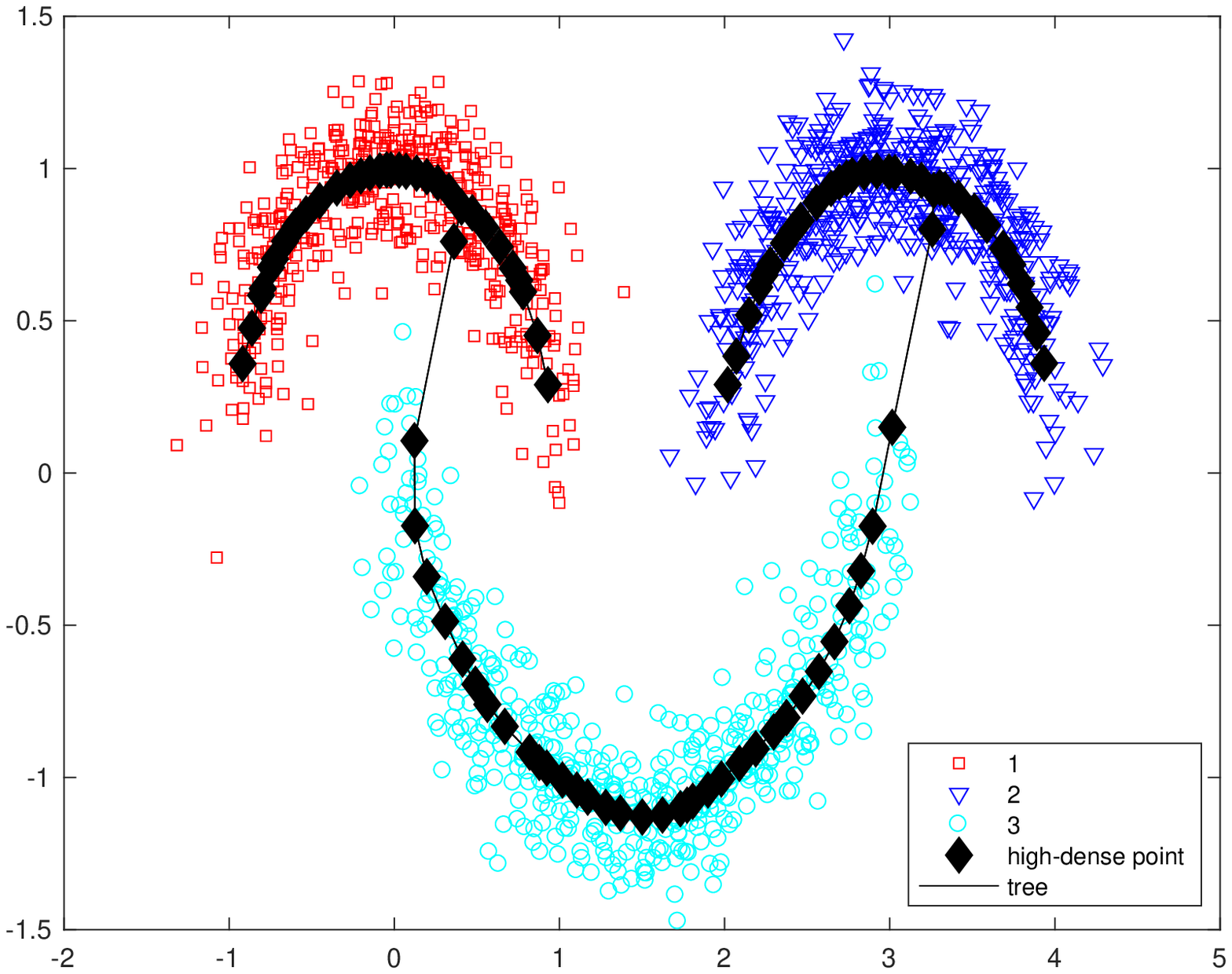} &
		\includegraphics[width=0.25\textwidth]{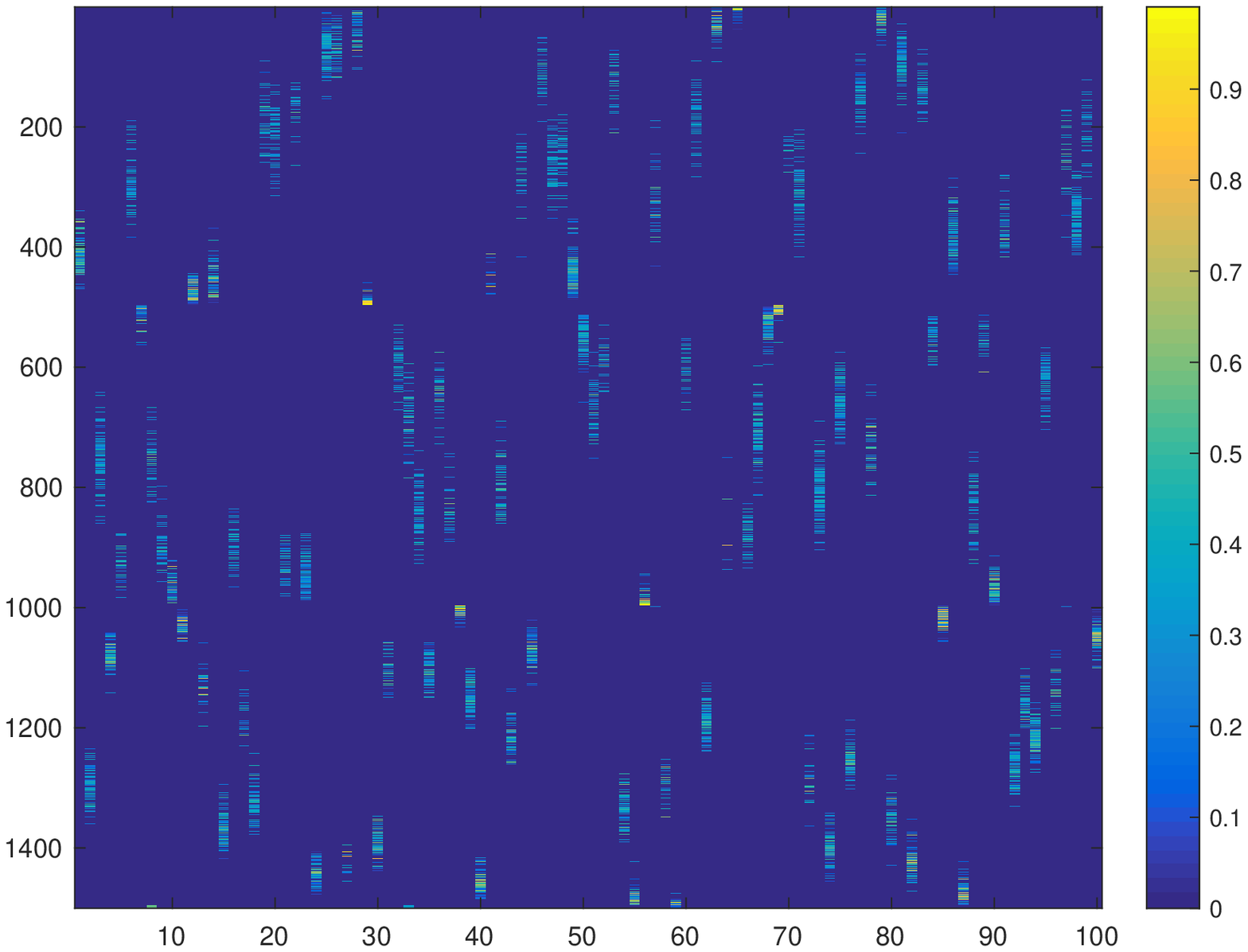}&
		\includegraphics[width=0.25\textwidth]{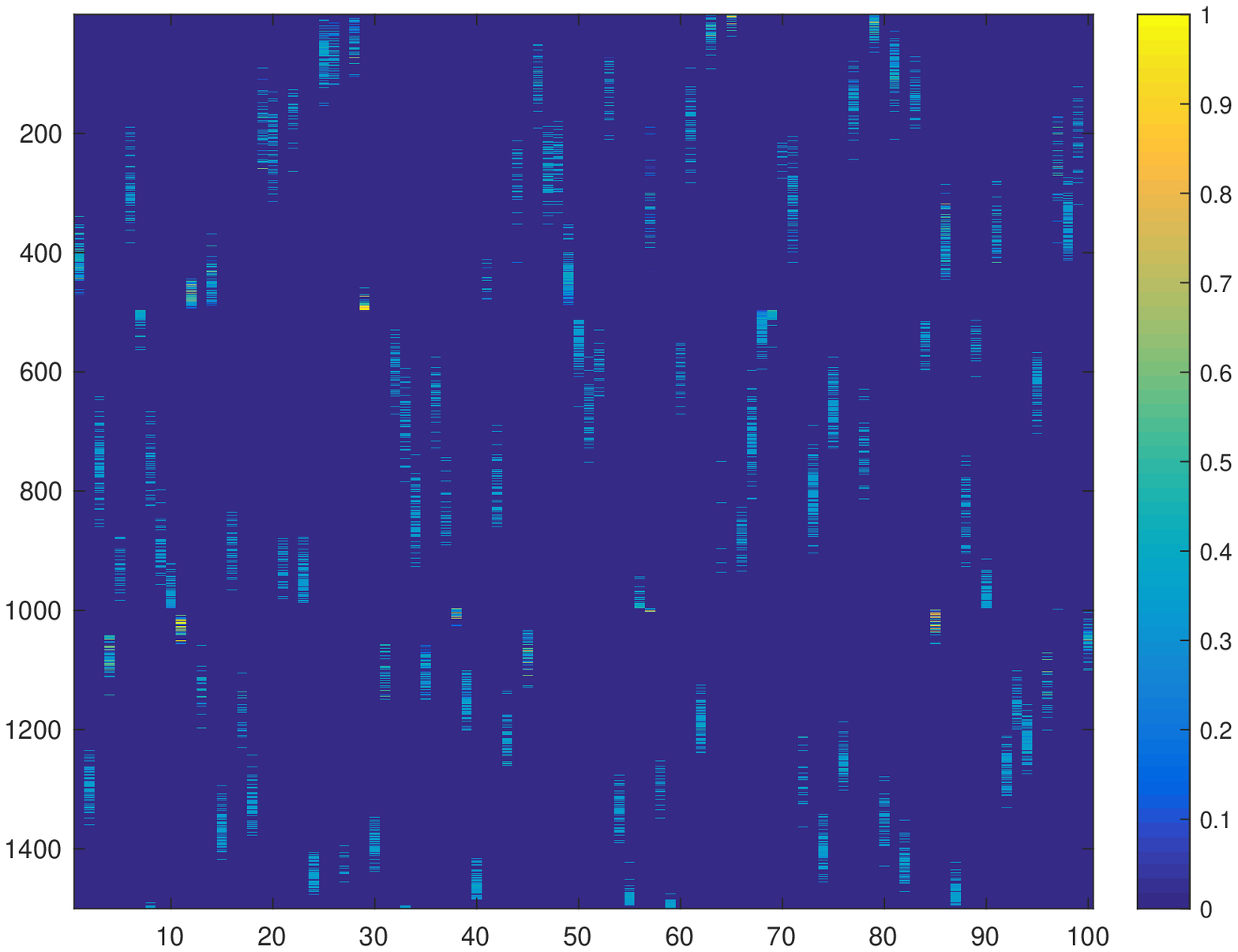}\\
		(e) anchor points & (f) $G$ over high-dense points & (g) $Z$ of AGR (LAE) & (h) $Z$ of AGR (Gauss)
	\end{tabular}
	\caption{The graph construction of three methods (LGC, AGR and our proposed method) on three-moon data. (a) the three-moon data points in 2-D space using the first two features. (b)-(c) the 10-NN graph and its affinity matrix used in LGC. (d) the $Z$ matrix used in our proposed method. (e) the anchor points obtained by the $k$-means method with $100$ centroids. (f) the optimized high-dense points and the learned tree structure. (g)-(h) the $Z$ matrices obtained by LAE  and the Nadaraya-Watson kernel regression with Gaussian kernel function in AGR, respectively.} \label{fig:three-moons-demo}
\end{figure*} 

\subsection{Graph Construction}
After obtaining $C$, $Z$ and $G$, we propose to construct the affinity matrix $W\in \mathbb{R}^{n\times n}$ by the following equation

\begin{small}\vspace{-0.15in}
	\begin{align}\label{con:Phi2}
	\begin{bmatrix}
	W & A_1\\
	A_2 & A_3
	\end{bmatrix}
	= P^2(I_{n+k}-\alpha P)^{-1},
	\end{align}
\end{small}\noindent
where $\alpha \in (0,1)$, and
\begin{small}
	\begin{align}\label{matrix:P:def}
	P & = \textbf{diag}\left( \begin{bmatrix}
	\textbf{0}_{n\times n} & Z\\
	Z^T &\eta G
	\end{bmatrix} \mathbf{1}_{n+k}\right)^{-1}  \begin{bmatrix}
	\textbf{0}_{n\times n} & Z\\
	Z^T &\eta G
	\end{bmatrix} \\
	& = \begin{bmatrix}  \textbf{0}_{n\times n} & \textbf{diag}(Z\mathbf{1}_k)^{-1}Z \\  \textbf{diag}(Z^T\mathbf{1}_n+\eta G\mathbf{1}_k )^{-1} Z^T & \eta  \textbf{diag}(Z^T\mathbf{1}_n+\eta G\mathbf{1}_k )^{-1} G \end{bmatrix} \nonumber \\
	&  =\begin{bmatrix} P_{11} & P_{12} \\
	P_{21} & P_{22}
	\end{bmatrix} = \begin{bmatrix} \textbf{0}_{n\times n}& Z \\
	P_{21} & P_{22}
	\end{bmatrix}. \nonumber
	\end{align}
\end{small}\noindent
$\textbf{0}_{n\times n} $ is the $n$ by $n$ zero matrix, $A_1, A_2$ and $A_3$ are dummy variables for representing $W$ in a compact form. $P_{11} =\textbf{0}_{n\times n}$, $P_{12} =Z$ since $Z\mathbf{1}_k = \mathbf{1}_n$. $\eta $ is a positive parameter to balance the scale difference between $Z$ and $G$. $Z$ is a positive matrix with $Z_{i,s}>0,~\forall i,s$ as defined in (\ref{eq:kde-Z}), and $G$ is a 0-1 matrix. The matrix inverse defined in (\ref{matrix:P:def}) always exists.  

Fig. \ref{fig:three-moons-demo} demonstrates three key differences of our graph construction approach from LGC and AGR on the synthetic three-moon data as shown in Table \ref{tab:datasets}: 1) the graph $W$ over all input data is implicitly represented by both $Z$ and $G$; 2) the high-dense points characterize the high-dense regions of the input data, instead of the simple centroids obtained by the $k$-means method; 3) the tree structure can effectively model the relationships among these high-dense points, while AGR does not have this property. Details of this synthetic data experiment can be found in Section IV-B.

For convenience of discussion, we denote 

\begin{small}\vspace{-0.1in}
	\begin{align}\label{def:Q}
	Q = \begin{bmatrix}
	\textbf{0}_{n\times n} & Z\\
	Z^T &\eta G  
	\end{bmatrix}, \quad E = \textbf{diag}(Z^T\mathbf{1}_n+\eta G\mathbf{1}_k ) 
	\end{align} 	
\end{small}\noindent
and 

\begin{small}\vspace{-0.1in}
	\begin{align}\label{def:Gamma}
	\Gamma = \textbf{diag}(Q\mathbf{1}_{n+k}) = \begin{bmatrix}
	I_n & 0\\
	0 & E
	\end{bmatrix},
	\end{align}	
\end{small}\noindent
then $P =  \Gamma^{-1}Q $ and $P\mathbf{1}_{n+k} = \mathbf{1}_{n+k}$, which satisfies the probability property over each row. We denote $X\geq 0$ if all elements in matrix $X$ are nonnegative. 

Before we prove matrix $W$ is symmetric, we would like to first show the following results:
\begin{prop} Let $P\in \mathbb{R}^{(n+k)\times (n+k)}$, and suppose $(I_{n+k}-\alpha P)^{-1} $ exists. Then $P^2(I_{n+k}-\alpha P)^{-1} = (I_{n+k}-\alpha P)^{-1} P^2$.
\end{prop}
\begin{proof}   It always holds:
	
	\begin{small} \vspace{-0.1in}
		\begin{align}\label{equal:twosides}
		P^2(I_{n+k}-\alpha P)  = (I_{n+k}-\alpha P) P^2= P^2 - \alpha P^3.
		\end{align} 
	\end{small}\noindent
	Since $(I_{n+k}-\alpha P)^{-1} $ exists, we multiply $(I_{n+k}-\alpha P)^{-1} $ on both sides of equation (\ref{equal:twosides}): 
	
	\begin{small}\vspace{-0.1in}
		\begin{align}\label{equal:twosides:2}
		(I_{n+k}-\alpha P)^{-1}P^2  =  P^2(I_{n+k}-\alpha P)^{-1}.
		\end{align} 	
	\end{small}\noindent
	The proof is completed.	
\end{proof}

\begin{lemma} Let $P =  \Gamma^{-1}Q $.  The eigenvalues of matrix $P$ are real, and lie in $[-1,1]$. 
\end{lemma}
\begin{proof} By the definition of matrix $P$, we have $P\geq 0$ and $P\mathbf{1}_{n+k} = \mathbf{1}_{n+k}$. So, all elements in matrix $P$ are between 0 and $1$. The row sums of matrix $P$ is 1. The characteristic equation of matrix $P$ is
	
	\begin{small}\vspace{-0.1in}
		\begin{align*}
		\det(\lambda I_{n+k} - P) = \det\left(\Gamma^{-\frac{1}{2}}\left( \lambda I_{n+k} -  \Gamma^{-\frac{1}{2}}Q\Gamma^{-\frac{1}{2}} \right) \Gamma^{\frac{1}{2}}\right).
		\end{align*}
	\end{small}\noindent	
	The	eigenvalues of matrix $P$ are the same as the eigenvalues of matrix $\Gamma^{-\frac{1}{2}}Q\Gamma^{-\frac{1}{2}} $.  Matrix $\Gamma^{-\frac{1}{2}}Q\Gamma^{-\frac{1}{2}} $ is symmetric, so it is eigenvalues are real, i.e., the eigenvalues of matrix $P$ are real. By Gershgorin circle theorem \cite{Gershgorin1931}, we have all the eigenvalues of matrix $P$ lie in $[-1,1]$.
\end{proof}

\begin{prop} Suppose $\alpha \in (0,1)$, $W$ defined in (\ref{con:Phi2}) is symmetric and nonnegative.
\end{prop}
\begin{proof} Since $Q = Q^T$, $P =  \Gamma^{-1}Q$, $P^T = Q \Gamma^{-1} = \Gamma P \Gamma^{-1} $.
	Since $P^2(I_{n+k}-\alpha P)^{-1} = (I_{n+k}-\alpha P)^{-1} P^2$,
	we have
	
	\begin{small}\vspace{-0.1in}
		\begin{align*}
		(P^2(I-\alpha P)^{-1})^{T} &= (I_{n+k}-\alpha P^{T})^{-1} (P^T)^2  \nonumber\\
		& = (I_{n+k} - \alpha \Gamma P \Gamma^{-1}  )^{-1} \Gamma P^2 \Gamma^{-1}  \nonumber\\
		& = \left(\Gamma (I_{n+k} - \alpha P) \Gamma^{-1}  \right)^{-1} \Gamma P^2 \Gamma^{-1}  \nonumber\\
		& = \Gamma (I_{n+k}- \alpha P) ^{-1}P^2\Gamma^{-1}  \nonumber\\
		& = \Gamma P^2 (I_{n+k} - \alpha P) ^{-1}\Gamma^{-1}.   
		\end{align*}
	\end{small}\noindent
	By the definition of matrix $P$, $\Gamma$ and $W$, we have left side: 
	
	\begin{small}\vspace{-0.1in}
		\begin{align}\label{left:side}
		(P^2(I-\alpha P)^{-1})^{T} =	\begin{bmatrix}
		W & A_1\\
		A_2 & A_3
		\end{bmatrix}^T = 	\begin{bmatrix}
		W^T & A_2^T\\
		A_1^T & A^T_3
		\end{bmatrix}.  
		\end{align} 
	\end{small}\noindent
	and the right side:   
	
	\begin{small}\vspace{-0.1in}
		\begin{align}\label{right:side}
		\Gamma P^2 (I - \alpha P) ^{-1}\Gamma^{-1}  
		&=  \begin{bmatrix}
		I_n & 0\\
		0 & E\end{bmatrix} 
		\begin{bmatrix}
		W & A_1\\
		A_2 & A_3
		\end{bmatrix} \begin{bmatrix}
		I_n & 0\\
		0 & E^{-1}\end{bmatrix} \nonumber \\
		& = \begin{bmatrix}
		W & A_1E^{-1}\\
		EA_2 & EA_3E^{-1}\end{bmatrix}. 
		\end{align}	  		
	\end{small}\noindent
	
	Comparing these two matrices in (\ref{left:side}) and (\ref{right:side}), we have
	$W = W^T$, i.e., matrix $W$ is symmetric. 
	
	Since the spectrum of matrix $P$ lies in $[-1,1]$, and $\alpha \in (0,1)$, we have
	$(I_{n+k}-\alpha P)^{-1} = \sum\limits_{t = 0}^{\infty} (\alpha P)^{t} $, i.e., the right side matrix series converge. Then
	
	\begin{small}\vspace{-0.1in}
		\begin{align}
		P^2(I_{n+k}-\alpha P)^{-1} = P^2 + \alpha P^3 +\alpha^2 P^4 + \cdots,
		\end{align}	
	\end{small}\noindent
	$P\geq 0$, so every term on the right hand side is nonnegative, and $W\geq 0.$ The proof is completed.
\end{proof}
By the proof of this proposition, we know matrix $W$ is symmetric, but the whole matrix $P^2(I-\alpha P)^{-1}$ defined in (\ref{con:Phi2}) is not necessarily symmetric. 

Next, we would like to show that the anchor graph defined in AGR is a special case of our proposed formulation (\ref{con:Phi2}). This is illustrated in Proposition \ref{prop:anchor-special-case}.

\begin{prop} \label{prop:anchor-special-case}
	Suppose $\widehat{Z}$ defined in (\ref{agr:Z}) is equal to $Z$ defined in (\ref{eq:kde-Z}). If $\eta = 0$ or $G = 0$ and $\alpha =0$, W defined in (\ref{con:Phi2}) is the same as anchor graph (\ref{eq:anchor_W}).
\end{prop}

Can we represent matrix $W$ explicitly? The answer is yes. We will show an explicit formula for matrix $W$. By the definition of matrix $P$ and matrix inversion in a $2$ by $2$ block form, we have the following equations:

\begin{small}\vspace{-0.1in}
	\begin{align}
	(I_{n+k}-\alpha P)^{-1} & =  \begin{bmatrix} I_n& -\alpha Z\\
	-\alpha E^{-1}Z^T & I_k-\alpha \eta E^{-1}G
	\end{bmatrix}^{-1} \nonumber\\
	& = \begin{bmatrix} L_{11}& L_{12} \\
	L_{21}& L_{22} \label{eq:IaPinv}
	\end{bmatrix},
	\end{align}	
\end{small}\noindent
where 

\begin{small}\vspace{-0.1in}
	\begin{align}
	L_{11} & = I_{n}+\alpha^2 Z(I_k-\alpha \eta E^{-1}G-\alpha^2 E^{-1}Z^TZ)^{-1}E^{-1}Z^T, \nonumber\\
	L_{12} & =  \alpha Z (I_k-\alpha \eta E^{-1}G-\alpha^2 E^{-1}Z^TZ)^{-1},\nonumber \\
	L_{21} & = \alpha (I_k-\alpha \eta E^{-1}G-\alpha^2 E^{-1}Z^TZ)^{-1} E^{-1}Z^{T}, \nonumber\\
	L_{22} & = (I_k-\alpha \eta E^{-1}G-\alpha^2 E^{-1}Z^TZ)^{-1}. \nonumber 
	\end{align}	
\end{small}\noindent
It is worth noting that (\ref{eq:IaPinv}) holds in the condition that matrix $I_k-\alpha \eta E^{-1}G-\alpha^2 E^{-1}Z^TZ$ must be invertible. The inversion is provided by the following lemma:   

\begin{lemma} \label{lemma:tildeP}
	For $\alpha \in (0,1)$, the eigenvalues of matrix 
	
	\begin{small}\vspace{-0.1in}
		\begin{align}\label{def:tildeP}
		\widetilde{P} = \alpha \eta E^{-1}G+\alpha^2 E^{-1}Z^TZ
		\end{align} 
	\end{small}\noindent	
	are real and lie in $(-1,1)$.
\end{lemma}

According to Lemma \ref{lemma:tildeP}, it is clear that $I_k-\widetilde{P}$ has eigenvalues in $(0, 2)$, so (\ref{eq:IaPinv}) holds for all $\alpha \in (0,1)$.
Accordingly, the right hand side of (\ref{con:Phi2}) is:

\begin{small}\vspace{-0.1in}
	\begin{align}
	P^2(I_{n+k}-\alpha P)^{-1}  =  \begin{bmatrix} W & A_1 \\ A_2 & A_3 \end{bmatrix},
	\end{align}	
\end{small}\noindent
where

\begin{small}\vspace{-0.1in}
	\begin{align}
	W & = ZE^{-1}Z^T L_{11}+\eta ZE^{-1}G L_{21}, \label{def:W:main}\\
	A_1 &= ZE^{-1}Z^TL_{12}+\eta ZE^{-1}GL_{22}, \nonumber\\
	A_2& = \eta E^{-1}GE^{-1}Z^T L_{11}+\left(E^{-1}Z^TZ + \eta^2(E^{-1}G)^2\right)L_{21} , \nonumber\\ 
	A_3 & = \eta E^{-1}GE^{-1}Z^T  L_{12}+\left(E^{-1}Z^TZ + \eta^2(E^{-1}G)^2\right) L_{22}. \nonumber
	\end{align}	
\end{small}\noindent
Substituting $L_{11}$ and $L_{21}$ into (\ref{def:W:main}) and simplifying, we get:

\begin{small}\vspace{-0.1in}
	\begin{align}\label{def:W:phi2}
	W = Z(I_k -\alpha \eta E^{-1}G-\alpha^2 E^{-1}Z^TZ)^{-1} E^{-1}Z^{T}.
	\end{align}	
\end{small}\noindent

Let us consider the affinity $W$ defined in (\ref{def:W:phi2}). Since $\widetilde{P}=\alpha \eta E^{-1}G+\alpha^2 E^{-1}Z^TZ$ has spectrum between $(-1,1)$, we have

\begin{small}\vspace{-0.1in}
	\begin{align}\label{appro:W}
	(I_k -\alpha \eta E^{-1}G-\alpha^2 E^{-1}Z^TZ)^{-1} =\sum\limits_{t=0}^{\infty} \widetilde{P}^t.
	\end{align}
\end{small}\noindent
If we only keep the first two terms, i.e., $t=0$ and $t= 1$, then we have an approximation of $W$ defined in (\ref{def:W:phi2}) as:

\begin{small}\vspace{-0.1in}
	\begin{align}
	\!\!\!\!&\widetilde{W}  = Z\left(I_k+\widetilde{P}\right)E^{-1}Z^{T} \nonumber \\
	\!\!\!\!& = Z\left(I_k+\alpha \eta E^{-1}G+\alpha^2 E^{-1}Z^TZ\right)E^{-1}Z^{T}  \nonumber \\
	& = Z E^{-1} Z^{T}\!+\!\alpha \eta ZE^{-1}GE^{-1}Z^{T}+\alpha^2 ZE^{-1}Z^TZE^{-1}Z^{T}  \label{def:W:phi1}   
	\end{align}
\end{small}\noindent
Obviously, matrix $\widetilde{W}$ is symmetric and nonnegative. Rather than storing $n\times n$ graph matrices $W$ and $\widetilde{W}$, we only need to store
the $n\times k$ matrix $Z$, $k\times k$ diagonal matrix $E$ and $k\times k$ matrix $Z^TZ$. We can easily use $Z,E,Z^TZ$ to get graph matrices (\ref{def:W:phi2}) and (\ref{def:W:phi1}). Our graph matrix constructions are very efficient for large-scale datasets. If $\alpha  = 0$, and $Z$ is given by solving (\ref{agr:Z}), $\widetilde{W}$ is the same as anchor graph $W$ that is defined in (\ref{eq:anchor_W}).

\subsection{Graph-based SSL} \label{sec:transductive}
Let $F = \begin{bmatrix}
F_l\\F_u
\end{bmatrix} \in \mathbb{R}^{n\times c}$ be the label matrix that mapping $n$ sample data $X$ to $c$ labels, where $F_l$ is a submatrix corresponding to samples with given labels and $F_u$ corresponds to unlabeled samples. We would like to learn the predicted label matrix $F_u$ by label propagation with the constructed graphs. 
Specifically, let \textbf{L} denotes the graph Laplacian operator, i.e.,

\begin{small}\vspace{-0.1in}
	\begin{align}
	\textbf{L}(W) = \textbf{diag}(W\mathbf{1}_n) - W，
	\end{align}	
\end{small}\noindent
where $W$ is either the exact $n\times n$ graph matrix defined in (\ref{def:W:phi2}) or the approximation graph matrix $\widetilde{W}$ defined in (\ref{def:W:phi1}). Next, we will show two different approaches to infer $F_u$ from $F_l$ and graph matrix $W$ or $\widetilde{W}$. 
 
\subsubsection{LGC-based approach}
By following the objective function of LGC \cite{Zhou:2003}, given a graph matrix $W$, we obtain the predicted label $F_u$ by solving the following optimization problem:

\begin{small}\vspace{-0.1in}
	\begin{equation}
	\min_{F_u\in\mathbb{R}^{(n-l)\times c}} \textbf{trace}(F^T\textbf{L}(W)F) + \frac{\lambda_2}{2} \|F-Y\|_{\text{fro}}^2
	\label{con: two-stage-obj2}
	\end{equation} 	
\end{small}\noindent
where   

\begin{small}\vspace{-0.1in}
	\begin{align}
	F = 
	\begin{bmatrix}
	F_l\\ 
	F_u	
	\end{bmatrix},
	\quad \text{and}\quad Y = \begin{bmatrix}
	Y_l\\
	Y_u
	\end{bmatrix}.	
	\end{align}
\end{small}\noindent
$\lambda_2>0$ is a regularization parameter. $Y_l$ is the assignment matrix for data points with known labels, and $F_l$ is the same as $Y_l$. So, $\|F-Y\|_{\text{fro}}^2 = \|F_u-Y_u\|_{\text{fro}}^2$. $Y_u$ follows a uniform initialization, with value $0$ at each of its entry. $F_u$ is the unknown soft-assignment matrix that we would like to solve. In equation (\ref{con: two-stage-obj2}), we do not impose any constraint on $F_u$, and we transform $F_u$ into an assignment matrix by selecting the maximum entry of each row. 

Problem (\ref{con: two-stage-obj2}) is an unconstrained quadratic programming with variable $F_{u}$. Let 

\begin{small}\vspace{-0.1in}
	\begin{align}\label{def:LW}
	\textbf{L}(W) = \begin{bmatrix}
	L_1(W) & L_2(W)\\ L_2^T(W) & L_3(W) 
	\end{bmatrix},
	\end{align}	
\end{small}\noindent
where $L_1(W)$ is the $l \times l$ block matrix, $L_2(W)$ is the $l\times (n-l)$ block matrix, $L_3(W)$ is the $(n-l)\times (n-l)$ matrix. The objective function of (\ref{con: two-stage-obj2}) is reformulated as:

\begin{small}\vspace{-0.1in}
	\begin{align}
	& \textbf{trace}(F^T\textbf{L}(W)F) + \frac{\lambda_2}{2} \|F-Y\|_{Fro}^2 \nonumber\\
	= &   \textbf{trace}\left(F_l^TL_1(W)F_l+2F_l^TL_2(W)F_u+F_u^TL_3(W)F_u\right)  \nonumber\\
	& \ +\frac{\lambda_2}{2}\|F_l-Y_l\|_{\text{fro}}^2 + \frac{\lambda_2}{2}\|F_u-Y_u\|_{\text{fro}}^2.
	\end{align}	
\end{small}\noindent

By taking derivative over $F_{u}$ and setting gradient over $F_{u}$ is 0, we get 

\begin{small}\vspace{-0.1in}
	\begin{equation}
	(2L_3(W) + \lambda_2 I_{n-l})F_u = \lambda_2 Y_u- 2L_2^T(W) F_l.
	\label{con: two-stage-Ft}
	\end{equation}	
\end{small}\noindent
Let $F_u = [F^1_u,\cdots, F_u^c]$, and $\lambda_2 Y_u- 2L_2^T(W) F_l = [b^1,\cdots,b^c]$. Problem (\ref{con: two-stage-Ft}) can be decomposed into $c$ linear equations that can be solved by conjugate gradient (CG) method \cite{numericaloptimization} in parallel: 

\begin{small}\vspace{-0.1in}
	\begin{equation}
	(2L_3(W) + \lambda_2 I_{n-l})F^i_u = b^i,\quad \forall i=1,\cdots,c.
	\label{con: two-stage-Ft-3}
	\end{equation} 
\end{small}\noindent
For very large number of classes, our method is very efficient. 

Let us recall an important result concerning the CG method. 
\begin{lemma}\emph{(\cite{li2008meinardus})}\label{lemma:CG}
	Let $A$ be a symmetric positive definite matrix with condition number $\kappa(A) \colon= \frac{\lambda_{\max}(A)}{\lambda_{\min}(A)}$, where $\lambda_{\max}, \lambda_{\min}$ denote the largest and smallest eigenvalues of matrix $A$ respectively. Then solving $Ax=b$ for $x$ using the CG method will converge in the following manner:
	
	\begin{small}\vspace{-0.1in}
		\begin{align}\label{cd:bound:1}
		\|x_t - A^{-1}b\|_A \leq 2 \Delta^{t}_{\kappa(A)}\|x_0 - A^{-1}b\|_A
		\end{align}		
	\end{small}\noindent
	where $\|v\|_A= \sqrt{v^TAv}$, 
	
	\begin{small}\vspace{-0.1in}
		\begin{align}\label{cd:bound}
		\Delta^{t}_{\kappa(A)} = \left(\left(\frac{\sqrt{\kappa(A)}+1}{\sqrt{\kappa(A)}-1}\right)^{t}+\left(\frac{\sqrt{\kappa(A)}-1}{\sqrt{\kappa(A)}+1}\right)^{t}\right)^{-1},
		\end{align} 
	\end{small}\noindent
	and $x_t$ is generated by CG at the $t$-th iteration, and $x_0$ is an initialization.
	\label{lemma: CG}
\end{lemma}
\begin{remark} $(\frac{\sqrt{\kappa(A)}-1}{\sqrt{\kappa(A)}+1})^t \rightarrow 0$ as $t\rightarrow \infty$ since $\frac{\sqrt{\kappa(A)}-1}{\sqrt{\kappa(A)}+1}<1$ for nonsingular $A$.  As
	
	\begin{small}\vspace{-0.1in}
		\begin{align}
		\Delta^{t}_{\kappa(A)}\leq \left(\frac{\sqrt{\kappa(A)}-1}{\sqrt{\kappa(A)}+1}\right)^t,
		\end{align}
	\end{small}\noindent	
	often, e.g. \cite{numericaloptimization}, inequality (\ref{cd:bound:1}) is weakened as:
	
	\begin{small}\vspace{-0.1in}
		\begin{align}\label{cd:bound:2}
		\|x_t - A^{-1}b\|_A \leq 2  \left(\frac{\sqrt{\kappa(A)}-1}{\sqrt{\kappa(A)}+1}\right)^t \|x_0 - A^{-1}b\|_A.
		\end{align} 	
	\end{small}\noindent
	
\end{remark}

Below, we would like to study the condition number of the coefficient matrix $2L_3(W) + \lambda_2 I_{n-l}$ in (\ref{con: two-stage-Ft-3}).
First, the eigenvalues of matrix $L_3(W)$ are shown in the following proposition: 
\begin{prop}\label{eig:bound:L3} 
	Suppose $\alpha \in (0,1)$. Let $W$ and $\widetilde{W}$ be the graph matrices defined in (\ref{def:W:phi2}) and (\ref{def:W:phi1}) respectively. Let $L_3(W)$ be the sub block matrix of $\textbf{L}(W)$ as defined in (\ref{def:LW}) with graph input matrix $W$ or $\widetilde{W}$. 
	\begin{itemize}
		\item For $W$, we have the row sums of matrix $W$ are in $[0, \frac{1}{1-\alpha}]$. The eigenvalues of $L_3(W)$ are real and lie in $[0, \frac{2}{1-\alpha}]$.
		\item For $\widetilde{W}$, we have the row sums of matrix $\widetilde{W}$ are in $[0,  1+\alpha]$. The eigenvalues of $L_3(\widetilde{W})$ are real and lie in $[0,  2(1+\alpha)]$.
	\end{itemize} 
\end{prop}

And then, we study the eigenvalues of coefficient matrix $2L_3(W) + \lambda_2 I_{n-l}$ in linear equations (\ref{con: two-stage-Ft-3}). Let 

\begin{small}\vspace{-0.1in}
	\begin{align}
	\Phi_1 & = 2L_3(W)+\lambda_2I_{n-l} \label{def:coef:W}\\
	\Phi_2 &= 2L_3(\widetilde{W})+\lambda_2I_{n-l} \label{def:coef:tidleW} 
	\end{align} 	
\end{small}\noindent
be the coefficient matrices of linear equations (\ref{con: two-stage-Ft}) with $W$ defined in (\ref{def:W:phi2}), and $\widetilde{W}$ defined in (\ref{def:W:phi1}). By Proposition \ref{eig:bound:L3}, we know the eigenvalues of $L_3(W)$ lie in $[0,\frac{2}{1-\alpha}]$, so the eigenvalues of coefficient matrix $\Phi_1 $ lie in $[\lambda_2,\lambda_2+\frac{4}{1-\alpha}]$. And, the eigenvalues of $L_3(\widetilde{W})$ lie in $[0,2(1+\alpha)]$, so the eigenvalues of coefficient matrix $\Phi_2$ lie in $[\lambda_2,\lambda_2+4(1+\alpha)]$.
The condition numbers of $\Phi_1$ and $\Phi_2$ are bounded by 

\begin{small}\vspace{-0.1in}
	\begin{align}
	\kappa(\Phi_1)&\leq \frac{\lambda_2+\frac{4}{1-\alpha}}{\lambda_2} = 1+\frac{4}{(1-\alpha)\lambda_2},\\
	\kappa(\Phi_2)&\leq \frac{\lambda_2+4(1+\alpha)}{\lambda_2} =  1+\frac{4(1+\alpha)}{\lambda_2}.
	\end{align}	
\end{small}\noindent
If parameters $\lambda_2$ and $\alpha$ are properly chosen (e.g., $1+\frac{4}{(1-\alpha)\lambda_2} < 10^{10}$), the coefficient matrices $\Phi_1$ and $\Phi_2$ will not be ill-conditioned. By Lemma \ref{lemma:CG}, the CG method will solve the linear equations (\ref{con: two-stage-Ft-3}) very efficiently. 

\subsubsection{AGR-based approach} Given a graph matrix $W$, we also consider to follow the label prediction procedure of AGR \cite{liu2010large} to do label prediction. Assume $F = ZA$ where $A\in \mathbb{R}^{k\times c}$, i.e., we represent the predicted label $F$ as a linear function of $Z$. And let $Z = \begin{bmatrix}
Z_{l} \\
Z_{u}
\end{bmatrix}$, we have 

\begin{small}\vspace{-0.1in}
	\begin{align}\label{con: two-stage-obj3}
	& \textbf{trace}(F^T\textbf{L}(W)F) + \frac{\lambda_2}{2} \|F-Y\|_{\text{fro}}^2 \nonumber\\
	= &~ \textbf{trace}(A^TZ^T\textbf{L}(W)ZA) + \frac{\lambda_2}{2} \|ZA-Y\|_{\text{fro}}^2 
	\end{align}	
\end{small}\noindent
If we minimize above function over variable $A$, we have

\begin{small}\vspace{-0.1in}
	\begin{align}\label{optimal:A}
	A^{*} = \lambda_2 (2Z^{T}\textbf{L}(W)Z+\lambda_2 Z^TZ)^{-1}(Z^T Y) 
	\end{align}	
\end{small}\noindent
After obtaining $A^*$, we can predict the unlabeled data as:
$F_{u} = Z_{u}A^*$, and choose the index of the maximum element in each row as the predicted label. The following terms can be computed efficiently without the need of any large matrix:
\begin{itemize}
	\item suppose graph matrix $W$ is defined in (\ref{def:W:phi2}), then 
	
	\begin{small}\vspace{-0.1in}
		\begin{align}
		Z^T\textbf{L}(W)Z &=  Z^{T}\left(\textbf{diag}(W\textbf{1}_n)-W\right)Z \nonumber \\
		& = Z^{T}\textbf{diag}(W\textbf{1}_n)Z -Z^{T}WZ \nonumber \\
		& = Z^{T}\textbf{diag}(Z(I_k -\widetilde{P})^{-1} E^{-1}Z^{T}\textbf{1}_n)Z \nonumber \\
		&\quad  -Z^{T}Z(I_k -\widetilde{P})^{-1} E^{-1}Z^{T}Z,
		\end{align}	
	\end{small}\noindent	
	which is a $k\times k$ matrix. 
	\item suppose graph matrix $\widetilde{W}$ is defined in (\ref{def:W:phi1}), then 
	
	\begin{small}\vspace{-0.1in}
		\begin{align}
		Z^T\textbf{L}(\widetilde{W})Z &=  Z^{T}\left(\textbf{diag}(\widetilde{W}\textbf{1}_n)-\widetilde{W}\right)Z \nonumber \\
		& = Z^{T}\textbf{diag}(\widetilde{W}\textbf{1}_n)Z -Z^{T}\widetilde{W}Z \nonumber \\
		& = Z^{T}\textbf{diag}(Z(I_k +\widetilde{P}) E^{-1}Z^{T}\textbf{1}_n)Z \nonumber \\
		&\quad  -Z^{T}Z(I_k +\widetilde{P})E^{-1}Z^{T}Z,
		\end{align}
	\end{small}\noindent	
	which is a $k\times k$ matrix. 
\end{itemize} 
As in (\ref{optimal:A}), the inverse is defined over a $k \times k$ matrix, so we can solve it directly using the inversion operation.

\subsection{Algorithm}
We have discussed the key components in Sections \ref{sec:high-dense} - \ref{sec:transductive}. By combining them, the proposed method is summarized in Algorithm \ref{alg:2} with two graph construction approaches and two inference approaches for the unlabeled data.

\begin{algorithm}
	\DontPrintSemicolon
	\KwInput{$k, \sigma, \lambda_1, \lambda_2, \alpha\in (0,1)$, and $\eta$}
	\KwOutput{$F, Z, C, G$}
	\KwData{$X, F_l = Y_{l}$}
	Initialization: uniform for $Y_u$, $k$-means for $C$, $Z$ by (\ref{eq:kde-Z})\;
	\While{not converge}{
		Solve $G$ using minimal spanning tree algorithm;\;
		$L = \textbf{diag}(G\textbf{1}_k) - G,\  \Xi  = \textbf{diag}(Z^T \textbf{1}_n)$; \; 
		$C \leftarrow X Z (\Xi + \lambda_1 L)^{-1}$;\;
		$Z_{i, s}\leftarrow\frac{\exp \left(-\left\|\mathbf{x}_{i}-\mathbf{c}_{s}\right\|^{2} / \sigma\right)}{\sum_{s=1}^{k} \exp \left(-\left\|\mathbf{x}_{i}-\mathbf{c}_{s}\right\|^{2} / \sigma\right)}, \forall i = 1,...,n, s=1,...,k$.
	} 
	Construct graph $W$ using either (\ref{def:W:phi2}) or (\ref{def:W:phi1})\\
	Update $F_u$ by solving (\ref{con: two-stage-Ft}) using CG method or (\ref{con: two-stage-obj3}) with the closed form solution $A^*$ in (\ref{optimal:A}).\;
	$ y_i = \arg\max\limits_{j\in\{1,\cdots,c\}}\{(F_u)_{i,j}\}, \forall i=l+1,\ldots,n$.
	\caption{High-dense graph learning (HiDeGL)}
	\label{alg:2}
\end{algorithm}

\section{Experiments} \label{sec:experiments}

\subsection{Experimental settings} \label{sec:settings}

We describe the experimental settings which are used in this paper to evaluate the proposed methods in Algorithm \ref{alg:2} by comparing with baselines on a variety of datasets. As discussed in Section \ref{sec:transductive}, our proposed methods are graph-based SSL methods based on two different proposed graph construction approaches in terms of (\ref{def:W:phi2}) and (\ref{def:W:phi1}), and the two inference approaches including LGC-based and AGR-based in Section \ref{sec:transductive}. In addition, AGR method \cite{liu2010large} extends LGC to handle large-scale data by either calculating Gaussian kernel regression (\ref{eq:Z-NWK}) or learning local anchor embedding (LAE) (\ref{agr:Z}) for graph construction as shown in Section \ref{sec:related-work}. Hence, we report the comparisons of our methods with LGC and AGR, together with several classical methods based on the following settings:

\begin{itemize}
	\item LGC \cite{Zhou:2003}. The affinity matrix is defined based on the $K$-NN graph with neighborhood parameter $K$ and Gaussian kernel with bandwidth parameter $\sigma$. LGC solves the regularization framework with a trade-off parameter $\mu$ to balance the smooth regularizer and the square loss function.
	
	\item AGR \cite{liu2010large}. Two large graph construction approaches are evaluated including the Nadaraya-Watson kernel regression based on the Gaussian kernel function and the $K$-NN graph, and the LAE by solving $n$ constrained optimization subproblems. The constructed graphs are then used for SSL with the regularization parameter $\gamma$. For the ease of reference, we name the AGR with the Nadaraya-Watson kernel regression as AGR(Gauss), and the AGR with LAE as AGR(LAE).
	
	\item Classical SSL methods are taken as the baselines for several benchmark datasets, including the $K$-NN classifier ($K$-NN), spectral graph transduction (SGT) \cite{joachims2003transductive}, Laplacian regularized least squares (LapRLS) \cite{belkin2006manifold}, $\mathcal{P}_{SQ}$ solved using SQ-Loss-1 \cite{subramanya2011semi}, and measure propagation (MP) \cite{subramanya2011semi}.
	
	\item TVRF \cite{TVRF}. TV-based multi-class graph partitioning with a region force is an approach to combine the graph cut in the spectral clustering method and a region force is inspired by the Chan-Vese model \cite{chan2001active}. The conditional probability of each vertex belonging to each class can be calculated in terms of the connectivity between two points through either one edge or two edges. We name the two approaches as TVRF(1) and TVRF(2), respectively.
	
	\item Our proposed high-dense graph learning approach (HiDeGL). It consists of two variants for graph construction from the input data: (\ref{def:W:phi2}) and (\ref{def:W:phi1}). LGC-based approach and AGR-based approach are employed by leveraging the learned graphs for SSL. Briefly, we name the LGC-based approach with graph matrix $W$ constructed by (\ref{def:W:phi2}) and (\ref{def:W:phi1}) as HiDeGL(L-accurate) and HiDeGL(L-approx) respectively. We name the AGR-based approach with graph matrix $W$ constructed by (\ref{def:W:phi2}) and (\ref{def:W:phi1}) as HiDeGL(A-accurate) and HiDeGL(A-approx) respectively. 
	
\end{itemize}

\begin{table}
	\caption{Datasets used in the experiments} \label{tab:datasets}
	\centering
	\vspace{-0.1in}
	\begin{tabular}{c|rcc}
		\hline
		Data Set & $n$ & $c$ & $d$\\
		\hline
		three-moon & 1,500 & 3 & 100 \\ 
		USPS-2 & 1,500 & 2 & 241\\
		COIL20 & 1,440 & 20 & 1024\\
		Opt-Digits & 5,620 & 10 & 64\\
		USPS & 9,298 & 10 & 256\\		
		Pendigits & 10,992 & 10 & 16\\
		Letter & 15,000 & 26 & 16\\		
		MNIST & 70,000 & 10 & 784\\	
		EMNIST-Digits & 280,000  & 10 & 784\\
		\hline
	\end{tabular}
	\vspace{-0.15in}
\end{table}

\begin{landscape}
\begin{table}[htbp]
	\caption{Average accuracies with standard deviations of nine methods over $10$ randomly drawn labeled data on three-moon data in terms of varied number of labels. Best results are in bold.} \label{tab:three-moon}
	\centering
	\vspace{-0.1in}
	\begin{small}
	\begin{tabular}{cccccccc}
		\hline
		method & $l$=3 & $l$=10 & $l$=25 & $l$=50 & $l$=75 & $l$=100 & $l$=150\\
		\hline 
		LGC & 94.19 $\pm$ 6.69 & 98.96 $\pm$ 0.49 & 99.02 $\pm$ 0.30 & 99.23 $\pm$ 0.13 & 99.29 $\pm$ 0.11 & 99.40 $\pm$ 0.12 & 99.34 $\pm$ 0.12\\ 
		TVRF(1) & 90.49 $\pm$ 4.80 & 97.48 $\pm$ 1.15 & 99.53 $\pm$ 0.03 & 99.52 $\pm$ 0.05 & 99.52 $\pm$ 0.04 & 99.56 $\pm$ 0.06 & 99.50 $\pm$ 0.06\\
		TVRF(2) & 99.52 $\pm$ 0.07 & 99.47 $\pm$ 0.09 & 99.46 $\pm$ 0.11 & 99.53 $\pm$ 0.03 & 99.54 $\pm$ 0.06 & 99.56 $\pm$ 0.06 & 99.52 $\pm$ 0.07\\
		AGR(Gauss) & 99.36 $\pm$ 0.32 & 99.46 $\pm$ 0.20 & 99.51 $\pm$ 0.25 & 99.65 $\pm$ 0.08 & 99.56 $\pm$ 0.22 & 99.61 $\pm$ 0.17 & 99.64 $\pm$ 0.12\\
		AGR(LAE) & 97.74 $\pm$ 1.41 & 98.68 $\pm$ 0.31 & 98.66 $\pm$ 0.39 & 98.83 $\pm$ 0.29 & 98.74 $\pm$ 0.45 & 98.76 $\pm$ 0.30 & 98.82 $\pm$ 0.36\\
		HiDeGL(L-approx) & 99.85 $\pm$ 0.06 & \textbf{99.86 $\pm$ 0.07} & \textbf{99.88 $\pm$ 0.06} & \textbf{99.88 $\pm$ 0.06} & \textbf{99.88 $\pm$ 0.05} & \textbf{99.90 $\pm$ 0.05} & \textbf{99.88 $\pm$ 0.06}\\
		HiDeGL(L-accurate) & 99.85 $\pm$ 0.05 & 99.85 $\pm$ 0.06 & \textbf{99.88 $\pm$ 0.05} & \textbf{99.88 $\pm$ 0.05} & \textbf{99.88 $\pm$ 0.05} & \textbf{99.90 $\pm$ 0.05} & 99.87 $\pm$ 0.06\\
		HiDeGL(A-approx)  & \textbf{99.87 $\pm$ 0.05} & \textbf{99.86 $\pm$ 0.07} & \textbf{99.88 $\pm$ 0.05} & \textbf{99.88 $\pm$ 0.06} & \textbf{99.88 $\pm$ 0.05} & 99.89 $\pm$ 0.06 & 99.87 $\pm$ 0.06\\
		HiDeGL(A-accurate) & 99.85 $\pm$ 0.09 & \textbf{99.86 $\pm$ 0.06} & 99.87 $\pm$ 0.05 & \textbf{99.88 $\pm$ 0.05} & \textbf{99.88 $\pm$ 0.05} & \textbf{99.90 $\pm$ 0.05} & 99.87 $\pm$ 0.06\\
		\hline
	\end{tabular}
\end{small}
\end{table}
\end{landscape}

The statistics of the datasets used in the experiments are shown in Table \ref{tab:datasets}. Two benchmark SSL datasets including USPS-2 and Opt-Digits are popularly used for evaluating the performance of SSL methods. 
USPS-2 is publicly available\footnote{http://olivier.chapelle.cc/ssl-book/benchmarks.html}, while Opt-Digits is downloaded from the UCI machine learning repository\footnote{archive.ics.uci.edu/ml/datasets/optical+recognition+of+handwritten+digits}. 
Moreover, the simulated three-moon data is used to demonstrate the detailed properties of HiDeGL. The data COIL20\footnote{http://www.cad.zju.edu.cn/home/dengcai/Data/MLData.html} is also used to show the performance of multi-class classification with a large number of classes.
To further illustrate the capability of HiDeGL for medium-size data, we conduct the experiments on datasets including USPS, Pendigits, Letter and MNIST, which are obtained from LIBSVM Data\footnote{https://www.csie.ntu.edu.tw/$\sim$cjlin/libsvmtools/datasets/}. The EMNIST-Digits provides balanced handwritten digit datasets directly compatible with the original MNIST dataset\footnote{https://www.nist.gov/itl/products-and-services/emnist-dataset} \cite{cohen2017emnist}.

Our experiments follow the settings in the work \cite{subramanya2011semi}. For graph-based SSL approaches, weighted graphs are defined on the $K$-NN graphs and Gaussian kernel where the number of neighborhood parameter $K$ and the bandwidth parameter $\sigma$ are tuned in terms of the mean accuracies over the 10 random drawn transduction sets. For the two benchmark datasets, we directly take the results from \cite{subramanya2011semi} under the same setting of the number of labeled data $l \in \{10, 50, 100, 150\}$. In addition, LGC ($\mu$), AGR ($\gamma$) and HiDeGL ($\lambda_2$) have the similar regularization parameter, which is tuned properly. Both AGR and HiDeGL take different ways to efficiently construct the affinity matrix for SSL. In AGR, $K$-NN graph is used, the affinity matrix is either constructed from the Gaussian kernel or by solving the LAE without further parameter, and the anchor points are obtained as the centroids of the $k$-means method. Further, the number of closest anchors are tuned in $[2,10]$ and $\gamma \in [0.001,1]$. In the proposed HiDeGL, given the number of high-dense points $k$, the high-dense points and the graph structure connecting these points are learned simultaneously based on parameters $\sigma$, $\lambda_1$. In addition, parameters $\alpha$ and $\eta$ are introduced in both (\ref{def:W:phi2}) and (\ref{def:W:phi1}) to achieve different properties of the constructed graph. As shown in Proposition \ref{prop:anchor-special-case}, the construction of AGR is a special case of HiDeGL in the case of $\eta=0$ or $G=0$ and $\alpha = 0$. In the experiments, we tune the parameters such as $k\in\{200, 500, 750,1500\}$, $\sigma\in [0.01, 0.5]$, $\lambda_1 \in \{0.1,1,10,100\}$, $\lambda_2 \in \{0.001, 0.01,0.05\}$, $\eta \in \{0.01, 0.1, 1\}$ and $\alpha \in [0.1,0.9]$. All these parameters are tuned based on the mean accuracies over the 10 random drawn transduction sets. The best mean accuracies and their standard deviations are reported.

\subsection{Synthetic data}
Three moon data consists of 1500 data points resided in $100$-dimensional space and equally divided into 3 classes. The dataset is generated as follows: 500 points in two-dimensional space are first randomly generated on an upper half circle centered at (1.5,0.4) with radius 1.5; and then, another 500 points in two-dimensional space are randomly generated on two lower half unit circle centered at (0,0) and (3,0) respectively; finally, the 1500 points in total are expanded to $100$ dimensions by filling up $98$ dimensions with zeros and adding noise following normal distribution with mean $0$ and standard deviation 0.14 to each of the 100 dimensions.

In Fig. \ref{fig:three-moons-demo}, we demonstrate the neighborhood graph structure with neighbor size equal to $10$ and its affinity matrix used in LGC, anchors in AGR with two approaches (Gauss and LAE) for obtaining $Z$, and our proposed graph construction approach by optimizing high-dense points and a tree structure. By comparing Fig. \ref{fig:three-moons-demo}(e) and Fig. \ref{fig:three-moons-demo}(f), it is clear to see the key difference between anchor points and high-dense points: 1) anchor points are the centroids obtained by the $k$-means method, while high-dense points locate in the high-dense regions, so they are different from cluster centroids; 2) the additional tree structure shown in Fig. \ref{fig:three-moons-demo}(f) is the unique feature comparing to the existing methods because it characterizes the skeleton structure of data and properly represents the similarities of high-dense points via the learned tree structure, where two high-dense points connected if they are similar. We notice that the matrices $Z$ obtained by AGR and HiDeGL are quite similar (see Fig. \ref{fig:three-moons-demo} (d), (g) and (h)). Hence, both methods are able to capture the relations between input data and representative points (anchor points in AGR and high-dense points in HiDeGL).

Table \ref{tab:three-moon} shows the average accuracies with standard deviations over $10$ randomly drawn labeled data obtained by the compared methods in terms of the varied number of labeled data. From Table \ref{tab:three-moon}, we have the following observations: 1) our proposed HiDeGL outperforms other methods over all varied number of labels; 2) For small number of labels such as $l \in \{ 3, 10 \}$, HiDeGL performs significantly better than others; 3) the four variants of HiDeGL with two graph construction approaches and two inference approaches for unlabeled data achieve almost similar accuracies. All these observations imply that our proposed methods are effective for SSL, especially for very small number of labeled data.

\begin{table}[t]
	\centering
	\caption{Average accuracies with standard deviations of compared methods over $10$ randomly drawn labeled data on three datasets in terms of varied number of labels. Best results are in bold.} \vspace{-0.1in}
	\label{tab:benchmark}
	\begin{small}
		\begin{tabular}{@{}l@{\hskip 0.1in}c@{\hskip 0.1in}c@{\hskip 0.1in}c@{\hskip 0.1in}c@{}}
			\hline
			Method & $l=10$ & $l=50$ & $l=100$ & $l=150$\\
			\hline\hline
			\multicolumn{5}{c}{USPS-2}\\			
			\hline
			k-NN	&	$80.0$	&	$90.7$	& 	$93.6$	& $94.9$\\
			SGT							&	$86.2$	&	$94.0$	&	$96.0$	&	$\mathbf{97.0}$\\
			LapRLS						&	$83.9$	&	$93.7$	&	$95.4$	&	$95.9$\\
			SQ-Loss-I					& 	$81.4$	&	$93.6$	&	$95.2$	&	$95.2$\\
			MP							&	$88.1$	&	$93.9$	&	$96.2$	&	$96.8$\\ 
			LGC & 85.21 $\pm$ 5.54 & 92.94 $\pm$ 3.36 & 95.94 $\pm$ 0.63 & 96.73 $\pm$ 0.28\\ 
			TVRF(1) & 82.00 $\pm$ 7.47 & 88.11 $\pm$ 2.85 & 92.47 $\pm$ 3.04 & 94.25 $\pm$ 1.80\\
			TVRF(2) & 73.66 $\pm$ 8.15 & 87.45 $\pm$ 4.19 & 92.86 $\pm$ 1.67 & 94.67 $\pm$ 1.05\\
			AGR(Gauss) & 75.01 $\pm$ 6.55 & 88.88 $\pm$ 2.65 & 91.92 $\pm$ 1.86 & 93.04 $\pm$ 1.04\\
			AGR(LAE) & 74.02 $\pm$ 8.60 & 88.01 $\pm$ 2.15 & 91.44 $\pm$ 1.39 & 92.33 $\pm$ 1.01\\
			HiDeGL(L-approx) & 90.01 $\pm$ 3.94 & \textbf{95.88 $\pm$ 0.50} & \textbf{96.23 $\pm$ 0.43} & 96.77 $\pm$ 0.39\\
			HiDeGL(L-accurate) & 89.41 $\pm$ 1.64 & \textbf{95.88 $\pm$ 0.50} & \textbf{96.36 $\pm$ 0.71} & 96.95 $\pm$ 0.25\\
			HiDeGL(A-approx) & 91.93 $\pm$ 3.69 & 95.30 $\pm$ 0.79 & 95.68 $\pm$ 0.81 & 96.16 $\pm$ 0.53\\
			HiDeGL(A-accurate) & \textbf{91.94 $\pm$ 3.68} & 95.30 $\pm$ 0.79 & 95.68 $\pm$ 0.81 & 96.16 $\pm$ 0.53\\ 
			\hline  
			\multicolumn{5}{c}{Opt-Digits}\\
			\hline
			k-NN						&	$79.6$	&	$85.5$	& 	$92.0$	& $93.8$\\
			SGT							&	$90.4$	&	$91.4$	&	$97.4$	&	$97.4$\\
			LapRLS						&	$89.7$	&	$92.3$	&	$97.6$	&	$97.3$\\
			SQ-Loss-I					& 	$92.2$	&	$95.9$	&	$97.3$	&	$97.7$\\
			MP							&	$90.6$	&	$94.7$	&	$97.0$	&	$97.1$\\ 
			LGC & 86.69 $\pm$ 2.54 & 93.79 $\pm$ 2.30 & 96.77 $\pm$ 0.55 & 97.06 $\pm$ 0.28\\ 
			TVRF(1) & 85.96 $\pm$ 3.37 & 92.81 $\pm$ 2.09 & 94.62 $\pm$ 1.04 & 96.54 $\pm$ 2.33\\
			TVRF(2) & 84.18 $\pm$ 5.23 & 97.15 $\pm$ 1.66  & \textbf{97.94 $\pm$ 0.15} & \textbf{97.98 $\pm$ 0.18}\\
			AGR(Gauss) & 60.40 $\pm$ 5.70 & 94.12 $\pm$ 3.38 & 96.15 $\pm$ 0.93 & 96.70 $\pm$ 0.50\\
			AGR(LAE) & 60.50 $\pm$ 5.33 & 94.07 $\pm$ 3.27 & 96.01 $\pm$ 1.07 & 96.56 $\pm$ 0.51\\
			HiDeGL(L-approx) & 92.16 $\pm$ 2.68 & \textbf{97.16 $\pm$ 0.91} & 97.55 $\pm$  0.60 & 97.90 $\pm$ 0.29\\
			HiDeGL(L-accurate) & 92.17 $\pm$ 2.69 & \textbf{97.16 $\pm$ 0.91} &97.55 $\pm$  0.60 & 97.89 $\pm$ 0.30\\
			HiDeGL(A-approx) & 93.65 $\pm$ 2.36 &  96.66 $\pm$ 0.87  & 97.12 $\pm$ 0.45 & 97.29 $\pm$ 0.33\\
			HiDeGL(A-accurate) & \textbf{93.69 $\pm$ 1.61} & 96.65 $\pm$ 0.86 & 97.12 $\pm$ 0.45 & 97.29 $\pm$ 0.32\\
			\hline   
			Method & $l=40$ & $l=80$& $l=100$ & $l=160$\\
			\hline
			\hline
			\multicolumn{5}{c}{COIL20}\\
			\hline 
			LGC & 87.39 $\pm$ 1.43 & 90.88 $\pm$ 1.53 & 93.43 $\pm$ 1.22 & 95.66 $\pm$ 1.13\\ 
			TVRF(1) & 89.31 $\pm$ 2.13 & 92.65 $\pm$ 0.92 & 94.24 $\pm$ 1.47 & 95.20 $\pm$ 1.06\\
			TVRF(2) & 87.19 $\pm$ 2.23 & 90.32 $\pm$ 2.33 & 92.42 $\pm$ 1.44 & 95.04 $\pm$ 0.74\\
			AGR(Gauss) & 84.16 $\pm$ 3.55 & 93.81 $\pm$ 2.20 & 94.09 $\pm$ 1.84 & 95.70 $\pm$ 1.51\\
			AGR(LAE) & 89.55 $\pm$ 3.22 & 97.19 $\pm$ 1.67 & 96.91 $\pm$ 1.73 & 98.24 $\pm$ 0.78\\
			HiDeGL(L-approx) & 92.95 $\pm$ 1.55 & 96.23 $\pm$ 0.88 & 96.37 $\pm$ 1.38 & 97.16 $\pm$ 1.70\\
			HiDeGL(L-accurate) & 91.20 $\pm$ 1.65 & 95.45 $\pm$ 1.30 & 96.37 $\pm$ 1.41 & 97.45 $\pm$ 0.77\\
			HiDeGL(A-approx) & \textbf{96.75 $\pm$ 1.51} & \textbf{97.88 $\pm$ 0.44} & \textbf{98.16 $\pm$ 0.94} & 98.58 $\pm$ 0.73\\
			HiDeGL(A-accurate)&  96.74 $\pm$ 1.43 & 98.04 $\pm$ 0.98 & 98.09 $\pm$ 0.74 & \textbf{98.66 $\pm$ 0.52}\\
			\hline		
		\end{tabular}
	\end{small}
\end{table}

\subsection{Classification performance}

We evaluate four variants of our proposed HiDeGL on varied sizes of datasets by comparing with baseline methods in terms of varied number of labeled data. The classification accuracy is used as the evaluation criterion. We repeat the experiments $10$ times by randomly drawing the given number of labeled data, and the average accuracies and their standard deviations are reported by tuning parameters of the corresponding methods as discussed in Section \ref{sec:settings}.

\begin{table}[t]
	\centering
	\caption{ Average accuracies with standard deviations of compared methods over $10$ randomly drawn labeled data on four medium-size datasets in terms of varied number of labels. Best results are in bold. Best results are in bold.} \label{tab:mediate-data}
	\vspace{-0.1in}
	\begin{small}
		\begin{tabular}{@{}l@{\hskip 0.1in}c@{\hskip 0.1in}c@{\hskip 0.1in}c@{\hskip 0.1in}c@{}}
			\hline
			Method	&	$l=10$ & $l=50$ & $l=100$ & $l=150$\\
			\hline
			\hline
			\multicolumn{5}{c}{MNIST}\\
			\hline		 
			LGC & 66.66 $\pm$ 5.52 & 83.76 $\pm$ 2.33 & 87.84 $\pm$ 1.11 &  89.41 $\pm$ 0.88\\
			TVRF(1) & 53.44 $\pm$ 6.73 & 74.35 $\pm$ 1.64 & 78.50 $\pm$ 1.70 & 81.27 $\pm$ 1.38\\
			TVRF(2) & 61.73 $\pm$ 6.12 & 78.05 $\pm$ 2.58 & 84.70 $\pm$ 1.20 & 86.19 $\pm$ 0.95\\
			AGR (Gauss) & 51.97 $\pm$ 4.15 & 76.05 $\pm$ 4.37 & 79.26 $\pm$ 0.68 & 80.32 $\pm$ 1.41\\
			AGR (LAE) & 52.29 $\pm$ 3.92 & 76.97 $\pm$ 4.37 & 80.33 $\pm$ 0.93 & 81.30 $\pm$ 1.45\\ 
			HiDeGL(L-approx)  & 75.69 $\pm$ 4.69&	\textbf{85.51 $\pm$ 1.94} &87.70 $\pm$ 0.66&89.48 $\pm$ 0.63\\ 
			HiDeGL(L-accurate)  & \textbf{75.73 $\pm$ 4.69} &\textbf{85.51 $\pm$ 1.95} & 87.72 $\pm$ 0.65&\textbf{89.50 $\pm$ 0.64} \\ 
			HiDeGL(A-approx) &  72.95 $\pm$ 3.83 & 85.22 $\pm$ 1.19 & 87.85 $\pm$ 1.08 & 89.08 $\pm$ 0.72\\
			HiDeGL(A-accurate) & 72.95 $\pm$ 3.83 & 85.21 $\pm$ 1.19 & \textbf{87.86 $\pm$ 1.07} & 89.08 $\pm$ 0.73\\ \hline
			\multicolumn{5}{c}{USPS}\\
			\hline		  
			LGC & 83.17 $\pm$ 5.24 & 93.90 $\pm$ 0.73 &  94.90 $\pm$ 0.30 &  95.04 $\pm$ 0.39\\
			TVRF(1) & 63.36 $\pm$ 6.90 & 81.87 $\pm$ 1.78 & 84.73 $\pm$ 0.95 & 85.58 $\pm$ 0.69\\
			TVRF(2) & 70.05 $\pm$ 7.52 & 81.65 $\pm$ 1.59 & 88.96 $\pm$ 0.19 & 88.90 $\pm$ 0.21\\
			AGR(Gauss) & 63.89 $\pm$ 10.30 & 93.35 $\pm$ 2.52 & 94.38 $\pm$ 0.48 & 94.77 $\pm$ 0.30\\
			AGR(LAE) & 63.51 $\pm$ 10.23 & 93.05 $\pm$ 2.39 & 94.31 $\pm$ 0.57 & 94.52 $\pm$ 0.36\\
			HiDeGL(L-approx) & 89.53 $\pm$ 5.46 & 94.96 $\pm$ 0.92 & 95.41 $\pm$ 0.26 & 95.55 $\pm$ 0.46\\
			HiDeGL(L-accurate) & 89.81 $\pm$ 4.80 & 94.96 $\pm$ 0.94 & 95.41 $\pm$ 0.26 & 95.53 $\pm$ 0.48\\
			HiDeGL(L-approx) & \textbf{91.42 $\pm$ 3.81} & 95.37 $\pm$ 0.36 & \textbf{95.50 $\pm$ 0.23} & \textbf{95.60 $\pm$ 0.25}\\
			HiDeGL(L-accurate)& 91.36 $\pm$ 3.86 & \textbf{95.38 $\pm$ 0.36} & 95.49 $\pm$ 0.22 & 95.59 $\pm$ 0.25\\
			\hline
			\multicolumn{5}{c}{Pendigits}\\
			\hline 
			LGC & 80.97 $\pm$ 7.41 & 93.21 $\pm$ 1.99 & 94.44 $\pm$ 1.39 & 95.89 $\pm$ 1.02\\
			TVRF(1) & 43.57 $\pm$ 4.20 & 59.52 $\pm$ 2.11 & 66.23 $\pm$ 2.57 & 74.69 $\pm$ 1.76\\
			TVRF(2) & 52.50 $\pm$ 4.05 & 83.39 $\pm$ 2.86 & 89.54 $\pm$ 2.80 & 92.99 $\pm$ 1.62\\
			AGR(Gauss) & 52.56 $\pm$ 6.85 & 91.73 $\pm$ 1.95 &  95.01 $\pm$ 1.03 & 96.43 $\pm$ 0.85\\
			AGR(LAE) & 52.52 $\pm$ 6.67 & 91.60 $\pm$ 1.88 & 94.59 $\pm$ 1.24 & 96.18 $\pm$ 1.21\\
			HiDeGL(L-approx) & 85.26 $\pm$ 4.09 & 93.36 $\pm$ 1.80 & 95.54 $\pm$ 1.00 &   \textbf{96.44 $\pm$ 1.06}\\ 
			HiDeGL(L-accurate) & \textbf{85.72 $\pm$ 4.08} & 93.24 $\pm$ 1.77  & \textbf{95.56 $\pm$ 0.91} &  96.36 $\pm$ 1.13\\
			HiDeGL(A-approx) &  83.01 $\pm$ 7.35 & \textbf{93.67 $\pm$ 2.00} & 95.44 $\pm$ 1.72 & 96.13 $\pm$ 0.86\\
			HiDeGL(A-accurate) & 82.89 $\pm$ 6.97 & \textbf{93.67 $\pm$ 2.00} & 95.44 $\pm$ 1.74 & 96.14 $\pm$ 0.87\\
			\hline 
			Method	&	$l=26$ & $l=52$ & $l=104$ & $l=156$\\
			\hline \hline
			\multicolumn{5}{c}{Letter}\\
			\hline 
			LGC & 31.12 $\pm$ 4.08 & 39.21 $\pm$ 2.54 & 52.46 $\pm$ 2.19 & 58.35 $\pm$ 2.14\\
			TVRF(1) & 19.49 $\pm$ 2.34 & 26.06 $\pm$ 2.85 & 33.23 $\pm$ 10.47 & 37.94 $\pm$ 11.99\\
			TVRF(2) & 22.73 $\pm$ 2.40 & 33.33 $\pm$ 4.08 & 45.79 $\pm$ 2.75 & 51.33 $\pm$ 1.47\\
			AGR(Gauss) & 25.44 $\pm$ 2.98 & 36.33 $\pm$ 2.31 & 49.18 $\pm$ 2.51 & 56.42 $\pm$ 1.99\\
			AGR(LAE) & 25.64 $\pm$ 2.68 & 36.42 $\pm$ 2.29 & 49.28 $\pm$ 2.55 & 56.43 $\pm$ 2.03\\
			HiDeGL(L-approx) & 31.89 $\pm$ 4.53 & 41.34 $\pm$ 2.73  & 53.32 $\pm$ 2.24&58.41 $\pm$ 1.29\\
			HiDeGL(L-accurate) & 32.02 $\pm$ 4.54 & 41.44 $\pm$ 2.81 &  53.17 $\pm$ 2.42  & 58.44 $\pm$ 1.35\\
			HiDeGL(A-approx)& 33.58 $\pm$ 4.43 & 42.02 $\pm$ 2.83 & \textbf{54.30 $\pm$ 1.97} & \textbf{58.97 $\pm$ 1.71}\\
			HiDeGL(A-accurate) & \textbf{33.59 $\pm$ 4.42} & \textbf{42.03 $\pm$ 2.80} &54.21 $\pm$ 1.34 & 58.57 $\pm$ 1.70\\
			\hline  
		\end{tabular}
	\end{small}
\end{table}

Our proposed methods can work efficiently for varied size of datasets, so we report results on benchmark datasets, medium-size data, and large-scale data, respectively. The average accuracies and their standard deviations are shown in Table \ref{tab:benchmark}, Table \ref{tab:mediate-data}, and Table \ref{tab:large-data:2}, respectively. Over all the sizes of tested datasets, HiDeGL gives the best accuracy than other methods for a small number of labeled data. On benchmark datasets, SGT is the best for $l=150$ on USPS-2 and TVRF(2) shows the best results for $l=100$ and $l=150$ on Opt-Digits. For medium-size data, the similar results can be observed for a small number of labeled data. For large number of labeled data, HiDeGL also shows better performance than others over all four datasets. For EMNIST-Digits, HiDeGL significantly outperforms AGR over all tested labels. These observations imply that our constructed graphs are effective for SSL.

We also demonstrate the running time of HiDeGL by comparing with AGR on EMNIST-Digits in terms of varied number of labeled data and $k=500$. Table \ref{tab:large-data:2} shows the CPU time of compared methods. Since the $k$-means method is used for both AGR and HiDeGL, we exclude the time for finding the anchor points or the initialization of Algorithm \ref{alg:2}. It is clear that 1) AGR(Gauss) is the fastest method, while its performance is the worst; 2) AGR(LAE) is the slowest since solving LAE for each point is time consuming; 3) HiDeGL with all four variants show the similar CPU time, and also demonstrates the best performance over all varied number of labels.

The above observations show that 1) HiDeGL works well for a small number of labels; 2) HiDeGL is scalable for large-scale data with a reasonable running time and good performance. Hence, our proposed graph construction approaches based on graph structure learning over high-dense points are effective and highly scalable for large-scale data with a small number of labeled data.

\begin{table}[t]
	\centering
	\caption{
		Average accuracies with standard deviations and CPUT time of compared methods over $10$ randomly drawn labeled data on EMNIST-digits in terms of varied number of labels. Best results are in bold.}	\label{tab:large-data:2}
	\vspace{-0.1in}
	\begin{small}
		\begin{tabular}{@{}l@{\hskip 0.08in}c@{\hskip 0.08in}c@{\hskip 0.08in}c@{\hskip 0.08in}c@{}}
			\hline
			Method	&	$l=10$ & $l=50$ & $l=100$ & $l=150$\\
			\hline
			\hline
			\multicolumn{5}{c}{Accuracy ($k=$ 500)}\\
			\hline				
			ARG(Gauss) & 47.25 $\pm$ 5.39 & 76.64 $\pm$ 1.22 & 80.41 $\pm$ 1.06 & 82.72 $\pm$ 0.98\\
			ARG(LAE) & 47.63 $\pm$ 5.07 & 78.06 $\pm$ 1.34 & 81.24 $\pm$ 0.85 & 83.32 $\pm$ 0.80\\
			HiDeGL(L-approx)  & 59.72 $\pm$ 8.34 & 80.63 $\pm$ 2.30 & \textbf{84.18 $\pm$ 1.56} & \textbf{85.37 $\pm$ 0.97}\\
			HiDeGL(L-accurate) & 60.74 $\pm$ 8.18 & 80.73 $\pm$ 1.74 & \textbf{84.18 $\pm$ 1.56} & \textbf{85.37 $\pm$ 0.97}\\
			HiDeGL(A-approx) & 67.58 $\pm$ 7.80 & \textbf{80.79 $\pm$ 1.88} & 84.04 $\pm$ 1.55 & 85.24 $\pm$ 0.95\\
			HiDeGL(A-accurate) & \textbf{67.69 $\pm$ 7.74} & \textbf{80.79 $\pm$ 1.88} & 84.04 $\pm$ 1.55 & 85.24 $\pm$ 0.95\\
			\hline
			\multicolumn{5}{c}{CPU Time (in seconds)}\\
			\hline		  	 
			AGR(Gauss) & 6.4 $\pm$0.44 & 6.75 $\pm$ 0.48 & 6.53 $\pm$ 0.58 & 6.41 $\pm$ 0.33\\  
			AGR(LAE)& 3430.1  $\pm$ 70.3 &  3409.9  $\pm$ 94.2 & 3368.5 $\pm$ 122.3 & 3391.9 $\pm$ 36.4 \\	
			HiDeGL(L-approx)& 242.7 $\pm$ 70.4 & 240.7  $\pm$ 43.9  & 229.7 $\pm$ 37.6 & 257.7 $\pm$ 44.0\\
			HiDeGL(L-accurate) & 	241.7   $\pm$ 70.4 &  239.8    $\pm$ 43.8 & 228.7 $\pm$ 37.6& 256.7 $\pm$ 43.9\\ 
			HiDeGL(A-approx) & 244.0 $\pm$ 70.5 & 243.8 $\pm$ 41.6 & 231.9 $\pm$ 36.3& 259.1 $\pm$ 44.4\\ 
			HiDeGL(A-accurate) & 	240.6  $\pm$ 70.4 & 240.4  $\pm$ 41.6& 228.7 $\pm$ 36.4& 255.4 $\pm$ 44.5\\ \hline
		\end{tabular}
	\end{small}
\end{table}
 
\subsection{Parameter sensitivity analysis}

In this subsection, we conduct the parameter sensitivity analysis of our proposed method using HiDeGL(L-accurate) as an illustration example in terms of different number of labeled data. Due to the multiple parameters, we perform the analysis for one parameter by fixing the others. Specifically, we report the best accuracy for the parameter over results obtained by tunning the others. The classification performance can be degraded if the number of high-dense points $k$ is too small, while it is time consuming if the number of high-dense points $k$ is large. To balance the classification performance and efficiency, we perform this analysis using $k=500$. The graph construction is the key of this paper, so we focus on the analysis of graph construction, and $\lambda_1, \sigma, \alpha$ and $\eta$ are studied. 

\begin{figure}
	\centering
	\begin{tabular}{@{}c@{}c@{}}
		\includegraphics[width=0.45\textwidth]{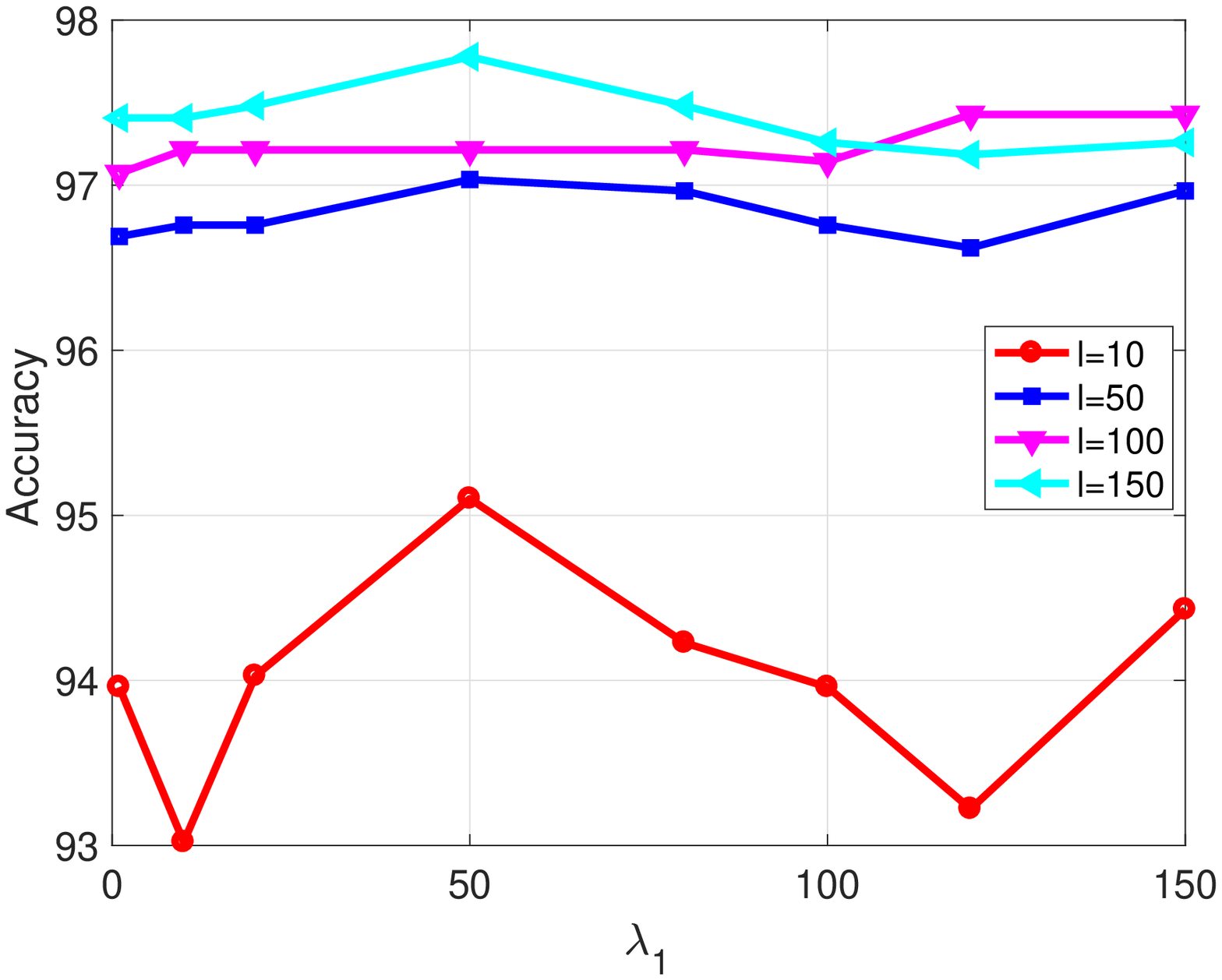} &
		\includegraphics[width=0.45\textwidth]{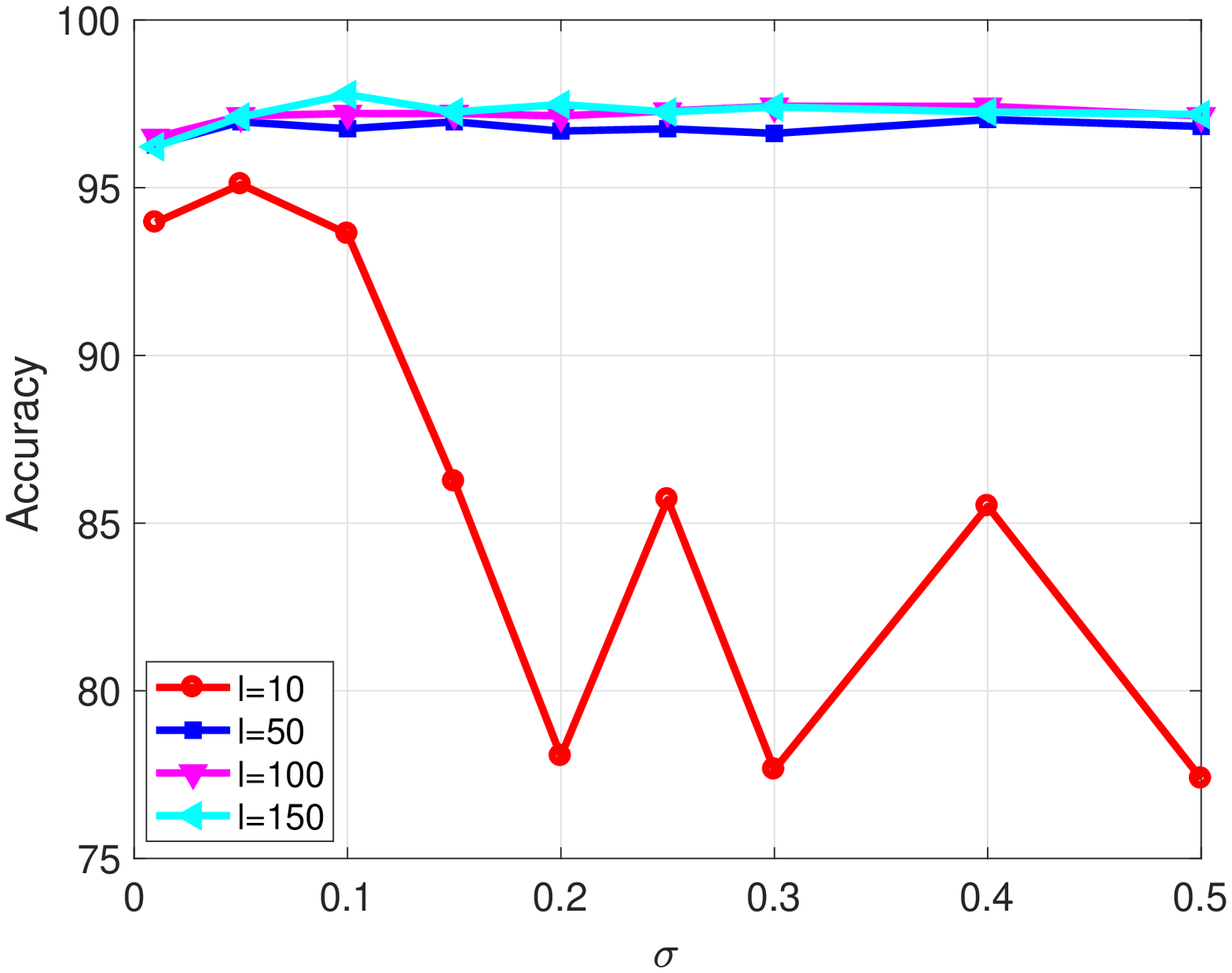}\\
		\includegraphics[width=0.45\textwidth]{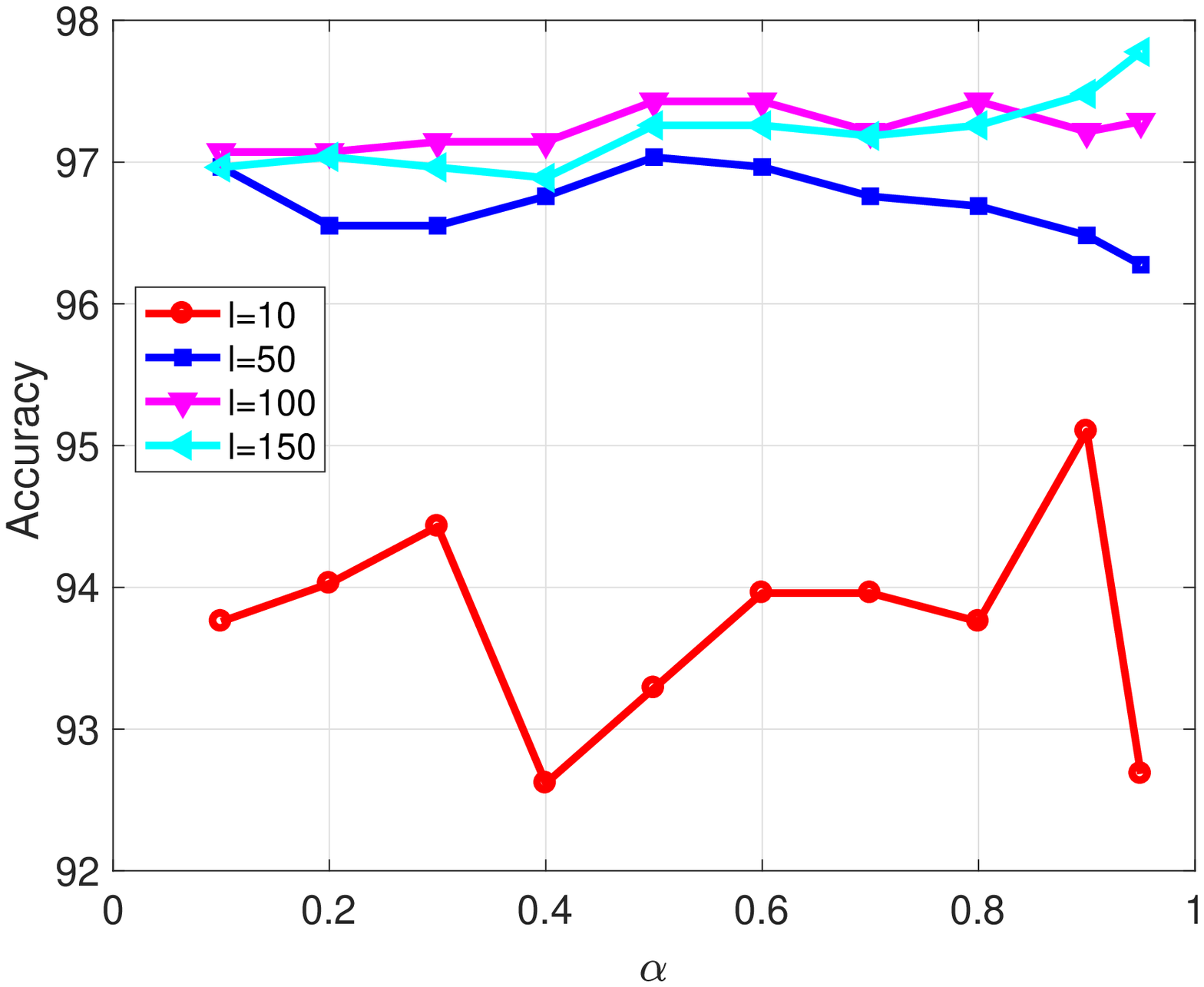} & 
		\includegraphics[width=0.45\textwidth]{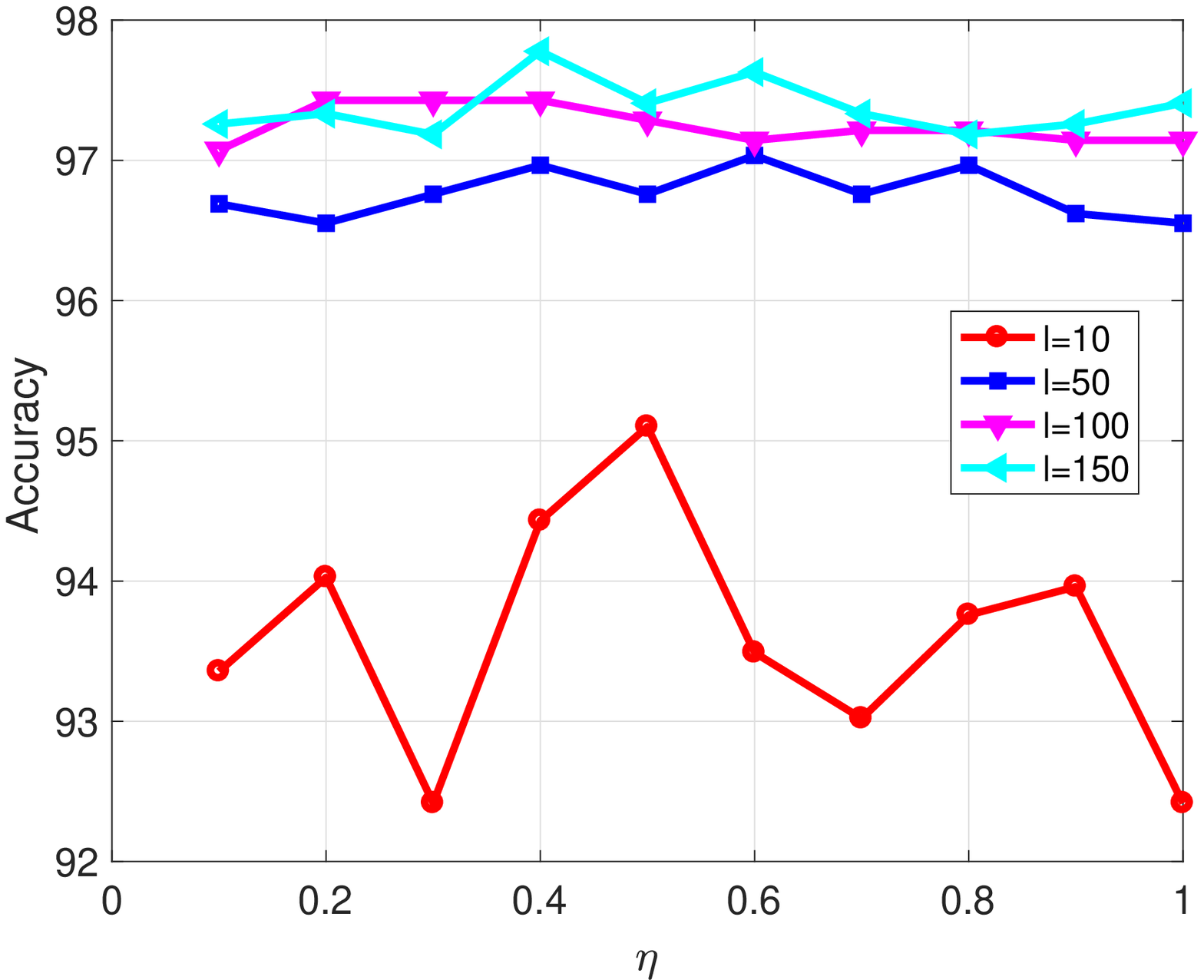}
	\end{tabular}
	\caption{Parameter sensitivity analysis of HiDeGL(L-accurate) on USPS-2 by varying the corresponding parameters $\lambda_1,\sigma,\alpha,\eta$ respectively with $k=500$ and $\lambda_2 \in \{10^{-3}, 10^{-2}\}$ in terms of the labeled set $l \in \{10, 50, 100, 150\}$.} \label{fig:param}
\end{figure}

Fig. \ref{fig:param} shows the accuracies of HiDeGL(L-accurate) by varying parameters in the given ranges and the number of labeled data. First, we notice that our method is quite robust for parameter $\lambda_1$ and $\eta$. Second, $\sigma$ and $\alpha$ can have large impact on the classification performance. It is clear to see that $\sigma$ changes more smoothly with a peak in $[0.05, 0.1]$. Since $\sigma$ controls the movement of points to its high-dense regions as shown in (\ref{eq:kde-Z}), the graph constructed over these points turns out to be important for better classification performance. Parameter $\alpha$ is also important since it is used for label propagation on a given graph by balancing the global and local assumptions as used in LGC. Third, the classification accuracies improve as the number of labeled data increases. However, the parameters are robust to the number of labeled data since the varying trends look quite similar.

\vspace{-0.1in}
\section{Conclusion} \label{sec:conclusions}
In this paper, we proposed a novel graph construction approach for graph-based SSL methods by learning a set of high-dense points, the assignment of each input data to these high-dense points, and the relationships over these high-dense points represented by a spanning tree structure. Our theoretical results showed various useful properties about the constructed graphs, and also AGR is a special case of our approach. Our experimental results showed that our methods not only achieved competitive performance to baseline methods but also were more efficient for large-scale data. More importantly, we found that our methods outperformed all baseline methods on the datasets with extremely small number of labels. 

\bibliographystyle{plain}
\bibliography{ssl_tree} 

\end{document}